# Semi-Supervised Learning with Heterophily

Wolfgang Gatterbauer
Carnegie Mellon University
gatt@cmu.edu

## ABSTRACT

We derive a family of linear inference algorithms that generalize existing graph-based label propagation algorithms by allowing them to propagate generalized assumptions about "attraction" or "compatibility" between classes of neighboring nodes (in particular those that involve heterophily between nodes where "opposites attract"). We thus call this formulation *Semi-Supervised Learning with Heterophily* (SSLH) and show how it generalizes and improves upon a recently proposed approach called Linearized Belief Propagation (LinBP). Importantly, our framework allows us to reduce the problem of estimating the relative compatibility between nodes from partially labeled graph to a simple optimization problem. The result is a very fast algorithm that – despite its simplicity – is surprisingly effective: we can classify unlabeled nodes within the same graph in the same time as LinBP but with a superior accuracy and despite our algorithm not knowing the relative compatibilities beforehand.

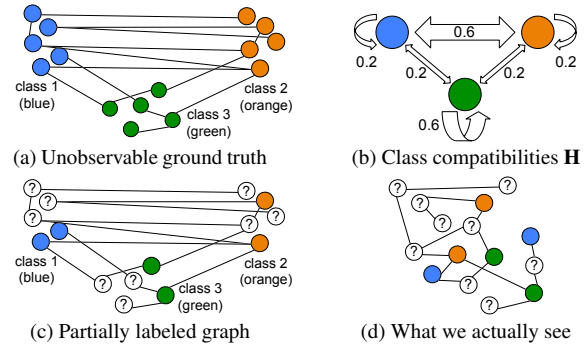

Figure 1: (a,b): Networks are formed based on relative affinities between the classes of nodes (here represented by their color). (c,d): The problem we are studying in this paper: We are given a *partially* labeled graph (i.e. the adjacency matrix **W** and a few classes of nodes) but we we do not know the relative affinities between classes. How can we *efficiently* (1) learn the relative affinities and (2) then predict the labels for the unlabeled nodes?

## 1. INTRODUCTION

An important problem in graph data management is *node labeling* (also called node classification). In a widely applicable scenario, we are given a graph with only part of the nodes labeled and where we assume that the (potentially weighted) edges reflect the strength of coupling between their end points. Given this input alone, *Semi-Supervised Learning* (SSL) methods attempt to infer the labels of the unlabeled data as closely as possible (fig. 2a).

Existing SSL methods commonly assume "smoothness" of labels over the graph, i.e. a certain *homophily* or *assortative mixing* property in the network that results in "birds of a feather flock together." For example, it is well-known that people with similar political affiliations are more likely to be linked or following each other in social networks.

However, the reverse is often true in actual data and is also called *heterophily* ("opposites attract"). For example, in English, nouns often follow adjectives, but seldom follow other nouns. A set of predators might form a function group in a biological food web, not because they eat each other, but because they eat similar prey (thus they "link" to similar nodes but not to each other) [6, 22]. In online auction settings, fraudsters are more likely linked to accomplices than amongst themselves [27]. Even some social networks have disassortative structure where pairs of nodes are more likely to be connected if they are from different classes (e.g., female and male members on dating websites). We are interested in these scenarios and even more general attractions among classes. For example, assume a social dating network with three different kinds of classes amongst its users. Class 1 prefers to date users of class 2 (and v.v.), whereas users of class 3 prefer to date among themselves (see fig. 1a).[1] We call these relations simply *heterophily relations*, as they naturally generalize notions of similarity/dissimilarity and assortative or disassortative mixing.[2]

In this paper, we show how a range of widely used SSL methods (e.g., those based on Local and Global Consistency (LGC) [42] or Harmonic Function methods (HF) [44]) can be generalized in a natural way to propagate heterophily from labeled to unlabeled data. We also show how we can learn the affinities between classes of neighbors from partially labeled data with a computationally remarkably simple framework.

Our first contribution relies on a simple but remarkable insight on the modulation properties of row-sum-constant matrices that are familiar from Markov chains:[3] Linearly transforming (or "modu-lating") a stochastic row vector **x** (which can represent the inferred label distribution of a node) with a row-stochastic matrix **H** results in a vector **f** that is still stochastic (**f** = **xH**). This property also holds in more general form: modulating a vector with row-sum $s$ with a row-stochastic matrix leads to another vector with row-sum $s$ (see fig. 2c). This property even holds when **H** and **x** have zero row-sums, i.e. when they are centered around 0. Thus, having influences from different neighbors, these influences can be superposed and still lead to a vectors centered around 0 (see appendix A for a more general discussion). This simple observation allows us to express a node's label distribution as a linear function of the label distributions of its neighbors and, furthermore, to express arbitrary similarity or dissimilarity relationships between neighboring nodes. We call the matrix **H** interchangeably potential or *compatibility* (modulation, affinity, coupling, heterophily) matrix, as it captures the pair-wise preference or coupling strengths between nodes and their classes in a network. This, in turn, allows us to derive a natural generalization of several well-known semi-supervised learning approaches by generalizing the commonly used regularization terms to situations with heterophily. We thus call our first contribution *Semi-Supervised Learning with Heterophily* (SSL-H).

---

[1]Notice that we imply undirected, i.e. symmetric relationships throughout this paper.
[2]In the physics community, homophily or assortative mixing is also referred to as "ferromagnetism," while heterophily or disassortative mixing is referred to as "antiferromagnetism."
[3]A Markov process is described by a state transition matrix **T** where $T_{i,j} = \mathbb{P}[X^{(t+1)} = i | X^{(t)} = j]$ is the probability to end up in state $i$ if starting at $j$ in the previous step. In that formalism, **T** is column stochastic and the column-wise sums are equal to 1: $\sum_i T_{i,j} = 1$ [5].

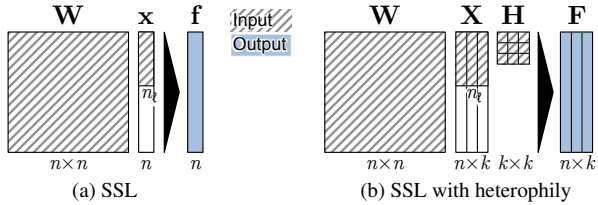
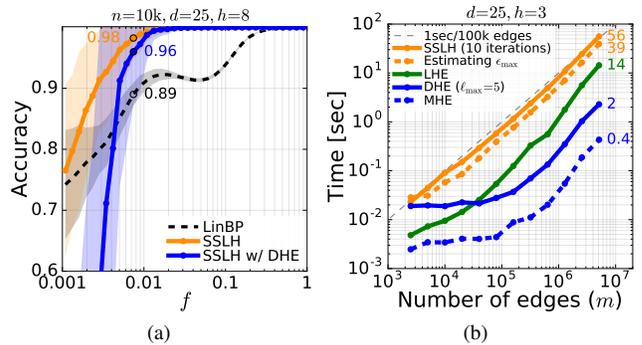

Figure 2: (a) In graph-based Semi-Supervised Learning (SSL), we infer a labeling function **f** for all $n$ nodes based on given labels **x** for $n_\ell < n$ nodes and the known graph structure **W** plus some "smoothness assumption" (also called "homophily," i.e. close nodes should have similar labels). (b) Our generalization allows arbitrary assumptions of relative coupling strengths between classes of neighboring nodes (e.g., "opposites attract"). (c) We express such arbitrary couplings with the help of a simple linear transformation by a symmetric and doubly stochastic compatibility matrix **H**. This paper shows not only (1) how to *infer* the missing labels from existing ones by using **H**, but also (2) how to also *learn* the compatibility matrix **H**.

Figure 3: (a): Our derived methods can considerably improve on the accuracy of existing methods, e.g., here 0.98 for "SSLH" instead of 0.89 for prior work "LinBP" for a graph with 10k nodes with only 80 labeled nodes. Our methods can even work better if the *compatibility between nodes is not given*, e.g., here still 0.96. (b): This additional step of estimating heterophily is extremely fast. Our most accurate method "DHE" learns the compatibility matrix on a graph with 5M edges in just 2 seconds.

The second problem is how to learn the affinities or compatibilities between labels from partially labeled data. Our second key contribution is more technical, and in the end a computationally simple way to learn the compatibility matrix **H**. Here, we draw an interesting connection to the Locally Linear Embedding (LLE) [29] framework. While the problems in LLE and our setup are different (LLE tries to find an optimal linear embedding across neighbors to reduce the dimensionality of the data; we find an optimal "heterophily explanation" between the classes of neighbors in a network), the mathematical formalism is similar. The resulting linear equations lead to a well-behaved optimization framework that can be solved very quickly with standard optimizers. Our second contribution is thus a family of algorithms for *Heterophily Estimation*. Together, LHE and SSL-H allow us to both (1) learn heterophily and to (2) infer the labels of unlabeled data (see fig. 2b).

**Contributions.** The two main contributions are thus:
1. *Semi-Supervised Learning with Heterophily (SSL-H)*: We generalize semi-supervised learning to general heterophily assumptions. The commonly used smoothness assumption is the important special case with the identify matrix as affinity matrix. The resulting framework can still be expressed by efficient matrix multiplications and implemented on top of existing linear algebra packages.
2. *Linear Heterophily Estimation*: We show how heterophily can be learned from existing partially labeled data even at the presence of few labels. This resolves the issue that the propagation matrix would otherwise have to be supplied by domain experts. The resulting framework is a simple convex optimization framework which can be solved with standard off-the-shelf solvers.

**Significance of contributions.** The resulting speed of our approach is significant fig. 3. We learn heterophily from graphs with millions of edges in less than 1min by using a standard optimizer in Python run on a single CPU. This is in stark contrast to common parameter estimation frameworks that rely on log-likelihood estimations and variants of expectation maximization which become computationally expensive for large datasets (e.g., a recent paper at KDD [22] develops a method that scales to graphs with maximal thousands of nodes). In order to achieve these speeds, our approach makes heavy use of *linear algebra*. Linear algebra is the key technical tool in machine learning that allows algorithms to scale, very similar to relational algebra in databases, and there are future paths to further increase the speed. For example, our results are achieved without yet basic optimization (we have not yet used gradient descent [15]). Even without that, and to the best of our knowledge, this is the first approach that can determine the relative compatibility (that are related to potentials in undirected graphical models) for graphs with millions of edges in 2 sec and with high accuracy on a single CPU hardware.

**Outline.** Section 2 gives extensive background on related work and lays the ground for our contributions. Section 3 derives an energy minimization framework that is requires for our later contributions. Section 4 shows how to generalize semi-supervised learning to heterophily. Section 5 shows how to learn heterophily from partially labeled data. Section 6 shows our first experiments before section 7 concludes.

|       |          | Update equation | Closed form | Energy function |
|-------|----------|-----------------|-------------|-----------------|
| binary | HF [44] | $\mathbf{f} \leftarrow \mathbf{x} + \bar{\mathbf{S}}\mathbf{W}^{\text{row}}\mathbf{f}$ | $\mathbf{f} = (\mathbf{I} - \bar{\mathbf{S}}\mathbf{W}^{\text{row}})^{-1}\mathbf{x}$ | $E(\mathbf{f}) = \mathbf{f}^\mathsf{T}\mathbf{L}\mathbf{f}$ s.t. $\mathbf{f}_\ell = \mathbf{x}_\ell$ |
|       | LNP [38] | $\mathbf{f} \leftarrow \bar{\alpha}\mathbf{x} + \alpha\mathbf{W}^{\text{row}}\mathbf{f}$ | $\mathbf{f} = \bar{\alpha}(\mathbf{I} - \alpha\mathbf{W}^{\text{row}})^{-1}\mathbf{x}$ | $E(\mathbf{f}) = \|\mathbf{f}-\mathbf{x}\|^2 + \mu\mathbf{f}^\mathsf{T}\mathbf{L}\mathbf{f}$ |
|       | LGC [42] | $\mathbf{f} \leftarrow \bar{\alpha}\mathbf{x} + \alpha\mathbf{W}^{\text{red}}\mathbf{f}$ | $\mathbf{f} = \bar{\alpha}(\mathbf{I} - \alpha\mathbf{W}^{\text{red}})^{-1}\mathbf{x}$ | $E(\mathbf{f}) = \|\mathbf{f}-\mathbf{x}\|^2 + \mu\mathbf{f}^\mathsf{T}\mathbf{L}^{\text{n}}\mathbf{f}$ |
|       | RWR [26] | $\mathbf{f} \leftarrow \bar{\alpha}\mathbf{x} + \alpha\mathbf{W}^{\text{col}}\mathbf{f}$ | $\mathbf{f} = \bar{\alpha}(\mathbf{I} - \alpha\mathbf{W}^{\text{col}})^{-1}\mathbf{x}$ | $E(\mathbf{f}) = \|\mathbf{f}-\mathbf{x}\|^2 + \mu\mathbf{f}^\mathsf{T}\mathbf{L}\mathbf{f}$ |
|       | FaBP [14] | $\mathbf{f} \leftarrow \mathbf{x} + h\mathbf{W}\mathbf{f} \underline{-h^2\mathbf{D}\mathbf{f}}$ | $\mathbf{f} = (\mathbf{I} - h\mathbf{W} \underline{+h^2\mathbf{D}})^{-1}\mathbf{x}$ | $E(\mathbf{f}) = \|\mathbf{f} - \mathbf{x} - h\mathbf{W}\mathbf{f} \underline{+h^2\mathbf{D}\mathbf{f}}\|^2$ |
| multi-class | HF | $\mathbf{F} \leftarrow \mathbf{X} + \bar{\mathbf{S}}\mathbf{W}^{\text{row}}\mathbf{F}$ | $\mathbf{F} = (\mathbf{I} - \bar{\mathbf{S}}\mathbf{W}^{\text{row}})^{-1}\mathbf{X}$ | $E(\mathbf{F}) = \sum_{i,j} W_{ij}\|\mathbf{F}_{i:} - \mathbf{F}_{j:}\|^2$ s.t. $\mathbf{F}_\ell = \mathbf{X}_\ell$ |
|       | LNP [38] | $\mathbf{F} \leftarrow \bar{\alpha}\mathbf{X} + \alpha\mathbf{W}^{\text{row}}\mathbf{F}$ | $\mathbf{F} = \bar{\alpha}(\mathbf{I} - \alpha\mathbf{W}^{\text{row}})^{-1}\mathbf{X}$ | $E(\mathbf{F}) = \|\mathbf{F}-\mathbf{X}\|^2 + \frac{\mu}{2}\sum_{i,j} W_{ij}\|\mathbf{F}_{i:} - \mathbf{F}_{j:}\|^2$ |
|       | LGC | $\mathbf{F} \leftarrow \bar{\alpha}\mathbf{X} + \alpha\mathbf{W}^{\text{red}}\mathbf{F}$ | $\mathbf{F} = \bar{\alpha}(\mathbf{I} - \alpha\mathbf{W}^{\text{red}})^{-1}\mathbf{X}$ | $E(\mathbf{F}) = \|\mathbf{F}-\mathbf{X}\|^2 + \frac{\mu}{2}\sum_{i,j} W_{ij}\|\frac{\mathbf{F}_i}{\sqrt{D_{ii}}} - \frac{\mathbf{F}_j}{\sqrt{D_{jj}}}\|^2$ |
|       | MRW [16] | $\mathbf{F} \leftarrow \bar{\alpha}\mathbf{U} + \alpha\mathbf{W}^{\text{col}}\mathbf{F}$ | $\mathbf{F} = \bar{\alpha}(\mathbf{I} - \alpha\mathbf{W}^{\text{col}})^{-1}\mathbf{X}$ | $E(\mathbf{F}) = \|\mathbf{F}-\mathbf{X}\|^2 + \frac{\mu}{2}\sum_{i,j} W_{ij}\|\mathbf{F}_{i:} - \mathbf{F}_{j:}\|^2$ |
|       | RNC [19] | $\mathbf{F} \leftarrow \mathbf{X} + \bar{\mathbf{S}}\mathbf{Z}^{-1}\mathbf{W}\mathbf{F}$ | not known | not known |
|       | LinBP [9] | $\mathbf{F} \leftarrow \mathbf{X} + \mathbf{W}\mathbf{F}\mathbf{H} \underline{-\mathbf{D}\mathbf{F}\mathbf{H}^2}$ | $\text{vec}(\mathbf{F}) = (\mathbf{I} - \mathbf{H}\otimes\mathbf{W} \underline{+\mathbf{H}^2\otimes\mathbf{D}})^{-1}\text{vec}(\mathbf{X})$ | $E(\mathbf{F}) = \|\mathbf{F} - \mathbf{X} - \mathbf{W}\mathbf{F}\mathbf{H} \underline{+\mathbf{D}\mathbf{F}\mathbf{H}^2}\|^2$ |

Figure 4: Comparison of several SSL methods for *inferring* the labels of unlabeled nodes. HF: harmonic function method [44], LNP: linear neighborhood propagation [38], LGC: local and global consistency method [42], RWR: random walks with restarts [26], MRW: MultiRankWalk [16], RNC: relational neighbor classifier [19], FaBP: fast belief propagation [14], LinBP: linearized belief propagation [9]. Underlined terms for FaBP and LinBP are optional "echo cancellation" (EC) terms; their highlighted energy functions are derived in section 3, and allow us later in section 5 to complement fast inference with similarly scalable *learning* of the compatibility matrix from partially labeled graphs.

## 2. FORMAL SETUP AND RELATED WORK

This section defines essential concepts while reviewing and comparing important related work on semi-supervised learning (section 2.1), random walks with restarts (section 2.2), belief propagation (section 2.3), and locally linear embedding (section 2.4). Since a main contributions of this paper is a generalization of various prior approaches, this section is longer than usual. For the same reason, we also use a unified notation that emphasizes similarities between different methods.

**General notations.** We make extensive use of matrix notation. We write both vectors ($\mathbf{x}$) and matrices ($\mathbf{X}$) in bold. We use row-wise ($\mathbf{X}_{i:}$), column-wise ($\mathbf{X}_{:j}$), and element-wise ($X_{ij}$) indexing of matrices. For example, $\mathbf{X}_{i:}$ is the $i$-th row vector of $\mathbf{X}$ (and therefore bold), whereas $X_{ij}$ is a single number (and therefore not bold).

### 2.1 Semi-Supervised Learning (SSL)

In graph-based *Semi-Supervised Learning* (SSL), we are given a graph of $n$ nodes among which $n_\ell$ nodes are labeled with a categorical class label from $L = \{1, \ldots, k\}$. WLOG all labeled nodes have indices $i \leq n_\ell$. The goal is to predict the class labels of the unlabeled nodes (i.e. for $n_\ell + 1 \leq i \leq n$) given the graph and various assumptions of "smoothness" between labels of neighboring nodes. Our following description of SSL combines the expositions and notations of several works [4, 17, 35, 40, 43, 45] in view of our later contribution. For the sake of consistency, our notation and formalism may at times differ from those used in original works.

Let $G = (V, E)$ be the graph with $n = |V|$, $m = |E|$, and real edge weights given by $w : E \to \mathbb{R}$. The weight $w(e)$ of an edge $e$ indicates the similarity of the incident nodes and a missing edge corresponds to zero similarity. The symmetric weighted *adjacency matrix* (or weight matrix) $\mathbf{W} \in \mathbb{R}^{n \times n}$ is defined by $W_{ij} \triangleq w(e)$ if $e = (i, j) \in E$, and 0 otherwise.[4] For the remainder of this paper, we assume that all edges have the same weight $w(e) = 1$. The diagonal degree matrix is defined as $\mathbf{D} \triangleq \text{diag}(D_{11}, \ldots, D_{nn})$ with $D_{ii} \triangleq \sum_j W_{ij}$. Throughout this paper, we assume a graph without any isolated vertices (thus $\mathbf{D}$ can be inverted). This is reasonable since we intend to infer labels of a node based on its neighbors, which becomes vacuous for an isolated node without neighbors.

Most binary SSL algorithms (e.g., [38, 42, 44]) specify the existing labels by a vector $\mathbf{x} = [x_1, \ldots, x_n]^\mathsf{T}$ with $x_i \in L = \{+1, -1\}$ for $i \leq n_\ell$ and $x_i = 0$ for $n_\ell + 1 \leq i \leq n$. Then a real-valued *labeling function* assigns a value $f_i$ with $1 \leq i \leq n$ to each data point $i$, based on which the final classification is performed as $\text{sign}(f_i)$, with $n_\ell + 1 \leq i \leq n$. In various algorithms, the values for the labeled nodes are either "*hard-clamped*" to the given ground labels and are never allowed to change, or "*soft-clamped*" and are thus allowed to differ from the given labels.[5]

The various SSL methods then differ in how they compute $f_i$ for each node $i$ and commonly justify their formalism from a "regularization framework", i.e. by motivating a different *energy function* (also varyingly called penalty, loss, or cost function) and proving that the derived labeling function $f$ is the solution to the objective of minimizing the energy function. The energy function commonly consists of two terms: (*i*) a *fit term* to existing labels, e.g. $(f_i - x_i)^2$ where $x_i$ denotes the given label and $f_i$ the learned label, and (*ii*) a *smoothness* or *regularization term*, e.g. $(f_i - f_j)^2$. The motivation is that if two points 1 and 2 are close (in a high-density region), then so should be the corresponding labels $f_1$ and $f_2$.

SSL methods are also sometimes called *transductive classification* [40]. In transductive inference, we reason from observed training cases directly to test cases in contrast to inductive inference, where we first infer general rules from training cases, and only then apply them to new test cases. This can be seen in that we are learning a labeling function that only applies to the given unlabeled nodes, and not a more general function. This distinction is important as transductive inference is generally seen as weaker (it only needs to predict labels for the supplied unlabeled nodes), but in turn, gives better predictions (it can focus on only the unlabeled nodes and not a more general labeling function).

We focus in the following on four influential methods: The first three (HF, LNP, LGC) were originally used for binary classification and are widely used in machine learning; the forth (RNC) was from the start targeted for mult-class classification and is more widely known in the data mining community.

**1) Harmonic Functions (HF).** The HF method [44] fixes the values of labeled nodes to their given ground labels and minimizes

---

[4]Notice that $|\mathbf{W}| = 2m$ since each undirected edge corresponds to two elements in $\mathbf{W}$.

[5]The notion "clamping" (hard and soft clamping) is used, e.g. in [23, Sec. 19] and by Scikit-learn [31]. An alternative notion is "soft label assignment"[13, 35].

the following energy function for the unlabeled nodes:

$$E(\mathbf{f}) = \frac{1}{2}\sum_{i,j} W_{ij}(f_i - f_j)^2$$

We can ensure hard-clamping by setting $\mathbf{f}_\ell = \mathbf{x}_\ell$ where $\mathbf{x}_\ell$ stands for the rows of $\mathbf{x}$ that correspond to the subset of indices of labeled nodes. We can then write the energy function in vector notation as

$$E(\mathbf{f}) = \sum_{i=1}^n f_i^2 \sum_{j=1}^n W_{ij} - \sum_{i,j=1}^n W_{ij}f_if_j = \mathbf{f}^\mathsf{T}(\mathbf{D}-\mathbf{W})\mathbf{f} = \mathbf{f}^\mathsf{T}\mathbf{L}\mathbf{f} \quad (1)$$

with the *unnormalized graph Laplacian*: $\mathbf{L} \triangleq \mathbf{D} - \mathbf{W}$.

The solution can be calculated iteratively by evaluating, for each unlabeled node, the new value as *weighted average value of its neighbors* (this is also called the "harmonic property"):

$$f_i \leftarrow \frac{\sum_{j=1}^n W_{ij}f_j}{D_{ii}}$$

These updates can also be written by defining the *row-stochastic adjacency matrix* $\mathbf{W}^{\text{row}} \triangleq \mathbf{D}^{-1}\mathbf{W}$ (thus $\sum_j W_{ij}^{\text{row}} = 1$), and $\mathbf{S}$ as the diagonal seed label indicator matrix with $S_{ii} = 1$ if $1 \leq i \leq n_\ell$, and 0 otherwise (i.e. $\mathbf{S}$ identifies the labeled nodes in the graph). Further define the diagonal identity matrix $\mathbf{I} \triangleq \text{diag}(1,\dots 1)$, where the dimensions are inferred from the context, and let $\bar{\mathbf{S}} \triangleq \mathbf{I} - \mathbf{S}$ (i.e. the diagonal identity matrix with ones present only for unlabeled nodes). Noticing that $\mathbf{S}\mathbf{x} = \mathbf{x}$, we can write:

$$\mathbf{f} \leftarrow \mathbf{x} + \bar{\mathbf{S}}\mathbf{W}^{\text{row}}\mathbf{f}$$

The closed-form is then: $\mathbf{f} = (\mathbf{I} - \bar{\mathbf{S}}\mathbf{W}^{\text{row}})^{-1}\mathbf{x}$.

**2) Linear Neighborhood Propagation (LNP).** The LNP method [38] is a label propagation method with one key difference to HF: The values for labeled nodes are soft-clamped and are thus allowed to differ from the given labels. In the iterative formulation, a node absorbs a fraction $\alpha$ ($0 < \alpha < 1$) of label information from its neighbors, and retains $\bar{\alpha} \triangleq 1 - \alpha$ of its initial state:

$$\mathbf{f} \leftarrow \bar{\alpha}\mathbf{x} + \alpha\mathbf{W}^{\text{row}}\mathbf{f}$$

In the first iteration, $\mathbf{f}^0 = \mathbf{x}$ on the right side. The closed form solution is then: $\mathbf{f} = \bar{\alpha}(\mathbf{I}_n - \alpha\mathbf{W}^{\text{row}})^{-1}\mathbf{x}$

Alternatively, LNP can be derived from a regularization framework by using the energy function:

$$E(\mathbf{f}) = \sum_{i=1}^n (f_i - x_i)^2 + \frac{\mu}{2}\sum_{i,j} W_{ij}(f_i - f_j)^2$$

This can be written in vector notation as:

$$E(\mathbf{f}) = ||\mathbf{f} - \mathbf{x}||_F^2 + \mu \mathbf{f}^\mathsf{T}\mathbf{L}\mathbf{f}$$

where $||\cdot||_F$ stands for the Frobenius norm of a vector or matrix (i.e. the square root of the sum of the squares of its elements). The update equation can then be recovered by substituting $\alpha = \frac{\mu}{2+\mu}$.

**3) Local and Global Consistency (LGC).** The LGC method [42] is similar to LNP in that given labels of nodes are soft-clamped. It is different to both above methods in that the labels are normalized by the square root of the degree $f_i/\sqrt{D_{ii}}$ when computing the similarity. The intended effect is that in graphs with highly varying degrees among nodes, the *influence of high degrees is reduced*.

The energy function for this method is given by:

$$E(\mathbf{f}) = \sum_{i=1}^n (f_i - x_i)^2 + \frac{\mu}{2}\sum_{i,j=1}^n W_{ij}\left(\frac{f_i}{\sqrt{D_{ii}}} - \frac{f_j}{\sqrt{D_{jj}}}\right)^2$$

The regularizer can be written more compactly as:

$$\frac{1}{2}\sum_{i,j=1}^n W_{ij}\left(\frac{f_i}{\sqrt{D_{ii}}} - \frac{f_j}{\sqrt{D_{jj}}}\right)^2 = \mathbf{f}^\mathsf{T}\mathbf{D}^{-\frac{1}{2}}(\mathbf{D}-\mathbf{W})\mathbf{D}^{-\frac{1}{2}}\mathbf{f} = \mathbf{f}^\mathsf{T}\mathbf{L}^{\text{n}}\mathbf{f}$$

with $\mathbf{L}^{\text{n}} \triangleq \mathbf{D}^{-\frac{1}{2}}\mathbf{L}\mathbf{D}^{-\frac{1}{2}} = \mathbf{I}_n - \mathbf{D}^{-\frac{1}{2}}\mathbf{W}\mathbf{D}^{-\frac{1}{2}}$ standing for the *normalized graph Laplacian*. By defining the *reduced adjacency matrix* $\mathbf{W}^{\text{red}} \triangleq \mathbf{D}^{-\frac{1}{2}}\mathbf{W}\mathbf{D}^{-\frac{1}{2}} = \mathbf{I}_n - \mathbf{L}^{\text{n}}$, the update equation becomes:

$$\mathbf{f} \leftarrow \bar{\alpha}\mathbf{x} + \alpha\mathbf{W}^{\text{red}}\mathbf{f}$$

and closed form: $\mathbf{f} = \bar{\alpha}(\mathbf{I}_n - \alpha\mathbf{W}^{\text{red}})^{-1}\mathbf{x}$.

**Observations.** The above three SSL algorithms show basically three ways of normalizing the adjacency matrix $\mathbf{W}$. The various normalizations represent different ways to help regularize the connection between closely-connected segments in a graph. Depending on assumptions of a particular graph problem, different regularizations may be advantageous (for example, to reduce the influence of nodes with high degrees). The three SSL algorithms can be formulated as (or are closely related) to the following energy function:

$$E(\mathbf{f}) = (\mathbf{f}-\mathbf{x})^\mathsf{T}\mathbf{S}(\mathbf{f}-\mathbf{x}) + \mathbf{f}^\mathsf{T}\mathbf{R}\mathbf{f} \quad (2)$$

where $\mathbf{R} \in \mathbb{R}^{n \times n}$ is the regularization matrix and $\mathbf{S}$ is a diagonal seed label indicator matrix with $S_{ii} = s_\ell > 0$ for $1 \leq i \leq n_\ell$, and $S_{ii} = s_u \geq 0$ for $n_\ell + 1 \leq i \leq n$, where $s_\ell$ and $s_u$ are two parameters [40]. The solution of eq. (2) can be shown to be

$$\mathbf{f} = (\mathbf{S}+\mathbf{R})^{-1}\mathbf{S}\mathbf{x}$$

For example, in HF the Laplacian matrix $\mathbf{L}$ is used for $\mathbf{R}$. And $s_\ell = \infty$, $s_u = 0$. That is, $f_i$ is hard-clamped and must be strictly equal to $x_i$ for $1 \leq i \leq n_\ell$, while there is no constraints on $f_i$ for $n_\ell + 1 \leq i \leq n$. For LGC, the normalized Laplacian matrix $\mathbf{L}^{\text{n}}$ is adopted for $\mathbf{R}$ and $0 < s_\ell = s_u < \infty$. Here $s_u = s_\ell$ arises from the label propagation process defined, and it has the effect of keeping $f_i$ of unlabeled $i$ within a reasonable range.

**Multi-class classification.** It is easy to extend existing label propagation algorithms to multi-class classification problems [38] by assigning a vector to each node, where each entry represents the belief of a node in a particular labeling class. Each of the classes is propagated separately and, at convergence, we adopt a "one-versus-all" approach. Suppose there are $k$ classes, and the label set is $L = \{1, 2, \dots, k\}$. Let $M$ be a set of $n \times k$ matrices with nonnegative real-valued entries. Any matrix $\mathbf{F} \in M$ corresponds to a specific classification on $X$ that labels node $i$ as $\text{argmax}_{c \leq k} F_{ic}$. Thus, $\mathbf{F}$ can be viewed as a $n \times k$-dimensional label matrix or, equivalently, as a function that assign labels for each node. The initial labels are given by an indicator matrix $\mathbf{X} \in \mathbb{R}^{n \times k}$: each row $i \leq n_\ell$ is described by $\mathbf{X}_{i:}$ with $X_{ic} = 1$ if node $i$ is labeled with class $c$ and $X_{ic} = 0$ otherwise. For each unlabeled node ($i > n_\ell$), $X_{ic} = 0$, for $(1 \leq j \leq k)$. Initially, we set $\mathbf{F}_0 = \mathbf{X}$. After convergence, the matrix $\mathbf{F}$ contains fitted label distributions (without necessarily probabilistic interpretation) with $F_{ic}$ giving the "belief" by which node $i$ believes in label $c$. Figure 4 shows the required changes to the energy function, update equations and closed-form solutions for HF, LNP, LGP, and RNC respectively (we will discuss FaBP and LinBP separately). Notice that labels for different classes *do not interact* with each other. In other words, they do not influence each other, and these models cannot model heterophily.

**Modeling dissimilarity and relation to this work.** Few papers have looked into ways to adapt the existing SSL frameworks to handle both similarity and dissimilarity. It is easy to see that simply encoding dissimilarity with negative edge weight is not appropriate as the energy function can become unbounded [43]. [10] defines a different graph energy function for dissimilarity edges. In particular, if $i$ and $j$ are dissimilar, one minimizes $W_{ij}(f_i + f_j)^2$. Note the essential difference to similarity edges is the plus sign instead of minus sign, and $W_{ij}$ stays non-negative. This forces $f_i$ and $f_j$ to have different signs and similar absolute values so they cancel each other out (the trivial solution of zeros is avoided by other similarity edges). The resulting energy function is still convex and can be easily solved using linear algebra. Similarly, [37] adopts a different objective function as minimizing a ratio, which is solved by a semidefinite program. Notice that both of these methods are limited to expressing binary dependencies between nodes (e.g., more similar or more dissimilar) and there is no obvious way to generalize these approaches to handle arbitrary heterophily. Even less clear is how the relative strengths of similarity and dissimilarity can be learned from existing data. This paper addresses these points.

## 2.2 Random walks with Restarts (RWR)

Similar to other semi-supervised learning methods, random walk-based methods make the assumption that the graph is homophilous, i.e. that instances belonging to the same class tend to link to each other or have higher edge weight between them [16]. In general, given a graph $G = (V, E)$, random walk algorithms return as output a ranking vector $\mathbf{f}$ that results from iterating following equation until convergence:

$$\mathbf{f} \leftarrow \bar{\alpha}\mathbf{u} + \alpha \mathbf{W}^{\text{col}}\mathbf{f} \qquad (3)$$

where $\mathbf{u}$ is a normalized teleportation vector with $|\mathbf{u}| = |V|$ and $||\mathbf{u}||_1 = 1$. Other than for $\mathbf{u}$, the formulation is equivalent to LNP [38] (see section 2.1) and the ranking vector $\mathbf{f}$ can be solved for analogously. Notice that above eq. (3) can be interpreted as the probability of a random walk on $G$ arriving at node $i$, with teleportation probability $\bar{\alpha}$ at every step to a node with distribution $\mathbf{u}$ [16]. Variants of this formulation are used by PageRank [25], Personalized PageRank [11, 7], Topic-sensitive PageRank [12], Random Walks with Restarts [26], and MultiRankWalk [16] which runs $k$ random walks in parallel (one for each class $c$).

We now give a formulation of MultiRankWalk [16] that was not stated before, and from which it will become clear later that it and other forms of rand walks are a special form of the framework developed in section 4: (1) For each class $c \in [k]$: (a) set $\mathbf{u}_i \leftarrow 1$ if node $i$ is labeled $c$, (b) normalize $\mathbf{u}$ s.t. $||\mathbf{u}||_1 = 1$. (2) Let $\mathbf{U}$ be the $n \times k$ matrix with column $i$ equal $\mathbf{u}_i$. (3) Then iterate until convergence:

$$\mathbf{F} \leftarrow \bar{\alpha}\mathbf{U} + \alpha \mathbf{W}^{\text{col}}\mathbf{F}\mathbf{I}_k$$

(4) After convergence, label each node $i$ with the class $c$ with maximum value: $c = \arg\max_j F_{ij}$.

Finally, notice that spectral clustering methods (e.g., spectral clustering by [24] and "*normalized cuts*" [34]) also have direct random walk interpretations. However, those methods are unsupervised and thus do not use any labeled data.

## 2.3 Belief Propagation (BP)

**Belief Propagation.** Another widely used method for semi-supervised reasoning in networked data is *Belief Propagation* (BP) [33] that has its origin in approximate probabilistic inference in undirected graphical models. Similar to the other SSL methods, BP helps to propagate the information from a few labeled nodes throughout the network by iteratively propagating information between neighboring nodes. In contrast to all prior methods, BP can also handle the case of heterophily, i.e. multiple classes influencing each other. However, BP has not closed-form, has no simple linear algebra formulation with matrix multiplication, and, importantly, for graphs with loops, BP has well-known convergence problems and despite a lot of work on convergence of BP (see e.g., [8, 21]) *exact* criteria for convergence are not known [23, Sec. 22]. Therefore, practical use of BP is still non-trivial (see [33] for a detailed discussion from a practitioner's point of view).

By using the symbol $\odot$ for the Hadamard product[6] and using $\mathbf{m}_{ji}$ for the $k$-dimensional *message* that node $j$ sends to node $i$, we can write the BP update equations [23, 39] for the belief vector of each node as:

$$\mathbf{f}_i \leftarrow Z_i^{-1} \mathbf{x}_i \odot \bigodot_{j \in N(i)} \mathbf{m}_{ji} \qquad \text{(BP)} \qquad (4)$$

Here, we write $Z_i$ for a normalizer that makes the elements of $\mathbf{f}_i$ sum up to 1. In parallel, each node sends messages to each of its neighbors:

$$\mathbf{m}_{ij} \leftarrow \mathbf{H}\Big(\mathbf{x}_i \odot \bigodot_{v \in N(i) \setminus j} \mathbf{m}_{vi}\Big) \qquad \text{(BP)} \qquad (5)$$

where $H_{cd}$ is a proportional "coupling weight" (or "modulation") that indicates the relative influence of class $c$ of a node on class $d$ of its neighbor. Thus, the outgoing messages are computed by multiplying together all incoming messages at a node (except the one sent by the recipient) and then passing through the $\mathbf{H}$ edge "potential." In view of our notation in this paper, we refer to this transformation by the edge potential as "modulation."

**Parameter learning.** The exact representation of probability distributions in graphical models, while very expressive from the modeling point of view, is statistically and computationally prohibitive. Typically learning of the potentials in a graphical model is done via maximum likelihood (ML) estimation i.e. finding the set of parameters that maximize the log-likelihood of the data, and iterative expectation-maximization (EM) methods [17, 32]. While in many cases these methods work well in practice, they often require careful initialization and problem-specific heuristics to yield good results. An additional challenge with our setup is that we estimate the parameters and the remaining labels on the same data set. For example, Sen and Getoor [32] consider a scenario in which they first learn a classifier on a fully labeled training data, and then apply the learned classifier to a separate partially labeled data set.

Our goal with this paper is to formulate the parameter estimation problem using efficient linear algebra calculations. We will later be able to cast the resulting estimation problem as a simple convex optimization problem while preserving full modeling power, thus providing a novel and very scalable approach to parameter estimation. For this, we build upon the following recent result:

**Linearized Belief Propagation.** [14] proposed to linearize belief propagation for the case of two classes and proposed fast belief propagation (FaBP) as a method to propagate existing knowledge of homophily or heterophily to unlabeled data. This framework is the result of a certain limit consideration of BP and allows to specify a "homophily factor" $h$ ($h > 0$ for homophily or $h < 0$ for heterophily) and to then use this algorithm with exact convergence criteria for binary classification (e.g., yes/no or male/female).

Recent follow-up work [9] derived a generalized solution for the multi-class case and called it "Linearized Belief Propagation"

---
[6]The Hadamard product (also called component-wise multiplication operator), is defined by: $\mathbf{Z} = \mathbf{X} \odot \mathbf{Y} \Leftrightarrow Z(i,j) = X(i,j) \cdot Y(i,j)$.

(LinBP). That work observed that the original update equations of BP can be reasonably approximated by linearized equations

$$\mathbf{f}_i \leftarrow \mathbf{x}_i + \frac{1}{k} \cdot \sum_{j \in N(i)} \mathbf{m}_{ji} \qquad \text{(LinBP)} \qquad (6)$$

$$\mathbf{m}_{ij} \leftarrow \mathbf{H}\Big(\mathbf{f}_i \underbrace{- \frac{1}{k}\mathbf{m}_{ji}}_{\text{EC}}\Big) \qquad \text{(LinBP)} \qquad (7)$$

by "*centering*" the belief vectors $\mathbf{x}, \mathbf{f}$ and the potential matrix around $\frac{1}{k}$. A vector or matrix $\mathbf{x}$ is centered around $c$ if all its entries are close to $c$ and their average is exactly $c$. If a vector $\mathbf{x}$ is centered around $c$, then the residual vector around $c$ is defined as $\hat{\mathbf{x}} = [x_1 - c, x_2 - c, \ldots]$ and centered around 0. In other words, the vectors become probability vectors and the potentials doubly stochastic matrices. Subsequently, this centering allows them to rewrite BP in terms of the residuals, and the derived messages remain centered for any iteration (thus no normalization is necessary).

The "echo cancellation" (EC) term is a result of the condition "$v \in N(i) \setminus j$" in the original BP equations and is mainly required for correspondence between BP and LinBP. We will thus distinguish between two versions of LinBP: the one with and the one without EC terms. Notice that no other linearized propagation framework considers such a term and we will separately investigate in the experiments the value of this term.

This formulation allows an iterative calculation of the final beliefs starting with an arbitrary initialization of $\mathbf{F}$ (e.g., all values zero):

$$\mathbf{F} \leftarrow \mathbf{X} + \mathbf{WFH}\underbrace{-\mathbf{DFH}^2}_{\text{EC}} \qquad (8)$$

This process was shown to converge if and only if the spectral radius $\rho$ of a matrix is smaller than 1:

$$\rho\Big(\mathbf{H} \otimes \mathbf{A} \underbrace{-\mathbf{H}^2 \otimes \mathbf{D}}_{\text{EC}}\Big) < 1 \qquad (9)$$

Here, $\otimes$ stands for the Kronecker product between two matrices. The closed form solution was shown to be:

$$\text{vec}(\hat{\mathbf{F}}) = \big(\mathbf{I}_{nk} - \mathbf{H} \otimes \mathbf{W} \underbrace{+ \mathbf{H}^2 \otimes \mathbf{D}}_{\text{EC}}\big)^{-1} \text{vec}(\mathbf{X}) \qquad (10)$$

Here, the *vectorization* $\text{vec}(\cdot)$ of a matrix stacks its columns one underneath the other to form a single column vector. That work also proposed to scale the matrix as

$$\mathbf{H}_\varepsilon = \varepsilon \mathbf{H} \qquad (11)$$

in order to guarantee convergence and has shown the quality of labeling to be widely independent of the "*scaling factor*" $\varepsilon$ and suggested a general default value of 0.5 of the value that would lead to divergence.

**Relation to this work.** Our paper builds upon the work on LinBP, yet generalizes and extends it in several ways: (*i*) We show how the previously shown SSL methods together with LinBP can be seen as particular instances in a wider design space of SSL that notably also includes "heterophily" in addition to the commonly used smoothness (= homophily) assumption (fig. 6). (*ii*) We derive a method to learn the compatibility matrix from partially labeled data. This avoids the reliance on domain experts. We achieve this by deriving the loss function that is minimized by LinBP (theorem 1). Importantly, our derivation leads to a simple convex optimization framework which allows us to deploy standard off-the-shelf optimizers for our purpose. (*iii*) We show that the labeling quality depends on the scaling factor $\varepsilon$ and, most interestingly, the highest labeling quality can be achieved above the exact phase transition between convergence and divergence.

## 2.4 Locally Linear Embedding (LLE)

Locally Linear Embedding (LLE, also sometimes referred to as Locally Linear Reconstruction) [29] is a method to derive compact representations of high-dimensional data by building a linear relationship among neighboring points. Our treatment here is brief; further details can be found in various sources [20, 29].

The LLE algorithm assumes the data consist of $n$ real-valued vectors $\mathbf{x}_i$, each of dimensionality $k$. In our notation, we write them as entries of a $n \times k$ matrix $\mathbf{X}$. LLE then reconstructs each data point from its neighbors by constructing a locally linear combination. Reconstruction errors are measured by the loss function

$$E(\mathbf{W}) = \sum_{i=1}^{n} ||\mathbf{X}_{i:} - \sum_{j=1}^{n} W_{ij}^{\text{row}} \mathbf{X}_{j:}||^2 \qquad (12)$$

which adds up the squared distances between all the data points and their reconstructions. The weights $W_{ij}^{\text{row}}$ summarize the contribution of the $j$-th data point to the $i$-th reconstruction. To compute the weights $W_{ij}^{\text{row}}$, one minimizes the loss function subject to the constraint of all rows of the weight matrix summing to one: $\sum_j W_{ij}^{\text{row}} = 1$. The optimal weights $W_{ij}^{\text{row}}$ are found by solving a least-squares problem and can be computed in closed form.

In the final step of LLE, each high-dimensional observation $\mathbf{X}_{i:}$ is mapped to a low-dimensional vector $\mathbf{F}_{i:}$ representing coordinates on the manifold. This is done by choosing $d < k$-dimensional coordinates $\mathbf{F}_{i:}$ to minimize the embedding loss function

$$E(\mathbf{F}) = \sum_{i=1}^{n} ||\mathbf{F}_{i:} - \sum_{j=1}^{n} W_{ij}^{\text{row}} \mathbf{F}_{j:}||^2 \qquad (13)$$

This loss function, like the previous one, is based on locally linear reconstruction errors, but here we fix the weights $W_{ij}^{\text{row}}$ while optimizing the coordinates $\mathbf{F}_{i:}$.

**Relation to this work.** LLE is originally an unsupervised learning algorithm. We will later use a similar formulation for estimating compatibility in partially labeled data, i.e. we will estimate the heterophily matrix $\mathbf{H}$ instead of the embedding $\mathbf{W}^{\text{row}}$ from data.

## 3. ALTERNATIVE ENERGY FUNCTIONS

The goal in this section is to formulate the solution to the update equations of LinBP [9] as the solution to an *optimization problem*, i.e. as an *energy minimization framework*. While the authors of LinBP derived the formulation from probabilistic principles (as approximation of the update equations of standard belief propagation), it is currently not known whether there is a simple objective function that a solution minimizes. Knowledge of such an objective is helpful as it allows principled extensions to the core algorithm. We will next derive the objective function for LinBP and will use it later in section 5 to solve the problem of *parameter learning*, i.e. learning the compatibility matrix from a partially labeled graph.

THEOREM 1 (LINBP LOSS FUNCTION). *The energy function minimized by the LinBP update equations is*

$$E(\mathbf{F}) = ||\mathbf{F} - \mathbf{X} - \mathbf{WFH}\underbrace{+\mathbf{DFH}^2}_{\text{EC}}||^2 \qquad (14)$$

**Discussion.** The corresponding energy function for FaBP in fig. 4 follows by substitution. We invite the reader to take a moment and compare the energy functions for LinBP with those of other multi-class SSL methods in fig. 4. Notice that, except for the EC term, the difference between the original weight matrix $\mathbf{W}$ and the reduced one $\mathbf{W}^{\text{red}}$, and the modulation by the compatibility matrix $\mathbf{H}$, the update equations of LinBP and LGC only differ by the factors $\alpha$ and $\bar{\alpha}$. However, the energy function has a fundamentally different form: LinBP has one single quadratic term, whereas LGC (and also all other SSL methods) have more than one quadratic term. As we will later see, the single quadratic term for LinBP will play a crucial role when we estimate the parameters from partially labeled data.

## 4. SSL WITH HETEROPHILY (SSL-H)

SSL algorithms rely on the "*fundamental smoothness assumption of semi-supervised learning*" [4, 35, 45]: *If two points* 1 *and* 2 *are close, then their respective labels* $f_1$ *and* $f_2$ *should be close.* To illustrate, consider the loss term $\mathbf{f}^\mathsf{T}\mathbf{R}\mathbf{f} \geq 0$ for the binary case where $\mathbf{R} \in \mathbb{R}^{n \times n}$ is the regularization matrix (e.g., $\mathbf{L}$ in the case of HF, recall eq. (1)). This term is also called "*smoothness term*" since it measures the smoothness of $\mathbf{f}$ on the graph (smaller values mean smoother $\mathbf{f}$). Roughly speaking, $\mathbf{f}$ is smooth if $f_i \approx f_j$ for those pairs with large $W_{ij}$. This property is sometimes informally expressed by saying that $f$ varies slowly over the graph, or that $\mathbf{f}$ "follows the data manifold" [45].

In the presence of heterophily (i.e. nodes of the same class being incompatible), this smoothness assumption does not hold anymore, for obvious reasons. Quite in contrary: for example, if in a dating network, females are more likely to connect to males, then the node labels will show the opposite of smoothness. We next present 6 ideas that generalize SSL to "*SSL with Heterophily (SSL-H):*"

**Idea 1: Modulated regularization.** Our first idea is to "modulate" the label distribution with a stochastic heterophily matrix $\mathbf{H}$ before sending it as message across an edge. We define our underlying assumption as follows:

DEFINITION 2 (SMOOTHNESS ASSUMPTION OF SSL-H). *If two points* 1 *and* 2 *are close, then the label vector* $\mathbf{f}_1$ *and the modulated label vector* $\mathbf{Hf}_2$ *should be close.*

Based on this assumption, we propose to replace the regularization term in multi-class SSL with the following term:

$$\mathbf{f}^\mathsf{T}\mathbf{R}(\mathbf{Hf}) \tag{15}$$

Here, $\mathbf{H}$ can either be a $[k \times k]$-dimensional symmetric doubly-stochastic matrix or the residual matrix after centering around $\frac{1}{k}$ (the equivalence of both versions is formally shown in appendix A). For example, one can generalize the approximate energy function for multi-class Linear Neighborhood Propagation (LNP, see fig. 4, and accordingly the update equations) to LNP with Heterophily (LNP-H) as:

$$E(\mathbf{F}) \approx ||\mathbf{F}-\mathbf{X}||^2 + \frac{\mu}{2}\sum_{i,j=1}^{n} W_{ij}||\mathbf{F}_{i:}-\mathbf{F}_{j:}\mathbf{H}||^2$$

Notice that the fundamental smoothness assumption of SSL corresponds to the special case in SSL-H of using the identity matrix $\mathbf{I}_k$ as idempotent modularizer, since $\mathbf{I}_k\mathbf{f} = \mathbf{f}$. We call our generalization of existing SSL algorithms "*modulated regularization.*"

**Idea 2: Energy function.** Our second idea originates from comparing the energy function for LinBP derived in section 3 with those of the other multi-class methods in fig. 4. Notice that after setting $\mathbf{H} = \mathbf{I}_k$, the update equations look similar except for relative factors $\alpha$ and $\bar{\alpha}$. Yet the energy function for LinBP consists of only a single quadratic term. We suggest to use this single quadratic term ("$||\cdot||^2$") for the energy function of SSL-H instead of the more common two terms, *fit term* and *regularization term* ("$\sum(\cdot)^2$"). The reason for this choice are twofold: (*i*) some of the update equations do not have a corresponding two term energy function (as proved in [36]). In contrast our energy minimization is always possible; (*ii*) the other reason will become apparent in section 5 where we estimate $\mathbf{H}$ from partially labeled data. It turns out that "$||\cdot||^2$" can be evaluated considerably faster than "$\sum(\cdot)^2$" since simple matrix multiplication can be performed efficiently. For example,

$$E(\mathbf{F}) = ||\mathbf{F}-\mathbf{X}-\mathbf{WFH}||^2$$

|       | soft clamping | hard clamping |
|-------|---------------|---------------|
| $\mathbf{W}$ | LinBP [9] $(0,0,0)$ | – |
| $\mathbf{W}^{\text{row}}$ | LNP [38] $(1,0,0)$ | HF [44] $(1,0,1)$ |
| $\mathbf{W}^{\text{red}}$ | LGC [42] $(\frac{1}{2},\frac{1}{2},0)$ | – |
| $\mathbf{W}^{\text{col}}$ | MRW [16] $(0,1,0)$ | – |

Figure 5: Various propagation matrices with either soft or hard clamping and their correspondences in our continous parameter space $(\alpha,\beta,\gamma)$.

**Idea 3: Hard clamping.** Our third idea is to model clamping by "*pushing*" clamping into the respective propagation matrix $\mathbf{W}^*$ before performing the update equations. We refer here with $\mathbf{W}^*$ to any of the four propagation matrices from fig. 4 or fig. 5. Recall from fig. 4 that we modeled seed labels with a diagonal seed label indicator matrix $\mathbf{S}$ with $S_{ii} = 1$ if $1 \leq i \leq n_\ell$, and 0 otherwise. Let's now redefine this matrix as diagonal "*clamping matrix*" $\mathbf{C}$ with $C_{ii} = 1$ if $i$ should be hard-clamped to its original value, and let $\bar{\mathbf{C}} \triangleq \mathbf{I} - \mathbf{C}$. Then $(\bar{\mathbf{C}}\mathbf{W}^*)$ is a modified propagation matrix that has all rows corresponding to clamped nodes removed. This corresponds to removing only *incoming* edges to clamped nodes, which means that those nodes will remain unmodified at each iteration.

$$E(\mathbf{F}) = ||\mathbf{F}-\mathbf{X}-(\bar{\mathbf{C}}\mathbf{W}^*)\mathbf{FH}||^2$$

**Idea 4: A continuum of propagation and clamping matrices.** Recall that $\mathbf{W}^{\text{row}} \triangleq \mathbf{D}^{-1}\mathbf{W}$, $\mathbf{W}^{\text{col}} \triangleq \mathbf{W}\mathbf{D}^{-1}$, and $\mathbf{W}^{\text{red}} \triangleq \mathbf{D}^{-\frac{1}{2}}\mathbf{W}\mathbf{D}^{-\frac{1}{2}}$. Notice that all four matrices from fig. 5 are just special cases of a family of propagation matrices defined by: $\mathbf{W}^{(\alpha,\beta)} \triangleq \mathbf{D}^{-\alpha}\mathbf{W}\mathbf{D}^{-\beta}$. We then recover our four matrices as $\mathbf{W} = \mathbf{W}^{(0,0)}$, $\mathbf{W}^{\text{row}} = \mathbf{W}^{(1,0)}$, $\mathbf{W}^{\text{col}} = \mathbf{W}^{(0,1)}$, and $\mathbf{W}^{\text{red}} = \mathbf{W}^{(0.5,0.5)}$. It is therefore natural to investigate the performance of SSL-H on ground truth data for all values of $(\alpha,\beta)$. Similarly, we can also model hard and soft clamping in a continuum by defining $\bar{\mathbf{C}}^{(\gamma)} \triangleq \mathbf{I} - \gamma\mathbf{C}$, and $0 \leq \gamma \leq 1$. We then recover soft clamping for $\bar{\mathbf{C}}^{(0)}$ and hard clamping for $\bar{\mathbf{C}}^{(1)}$. To see why various degree of clamping are an expressive addition to linearized propagation, notice that hard-clamping models multiplicative belief propagation with deterministic a prior beliefs. For example, an explicit belief $\mathbf{x} = (1,0,0)$ guarantees that the belief stays the same throughout future iterations. This can normally not be achieved in a linear system as a strong influence from a set of neighbors can always overwrite an explicit belief. Yet clamping (in various strengths) can compensate for that.[7] For notational conciseness, we combine the parameters and write $\mathbf{W}^{(\alpha,\beta,\gamma)} \triangleq \bar{\mathbf{C}}^{(\gamma)}\mathbf{W}^{(\alpha,\beta)}$. We thus have:

$$E(\mathbf{F}) = ||\mathbf{F}-\mathbf{X}-\mathbf{W}^{(\alpha,\beta,\gamma)}\mathbf{FH}||^2$$

**Idea 5: Echo cancellation.** Our fifth idea is to introduce the echo cancellation (EC) term from LinBP to other variants of SSL. This term is originally motivated based on sound probabilistic reasoning in the original BP update equations: a node sends a message across an edge that *excludes* information received across the same edge from the other direction ("$v \in N(i) \setminus j$" in eq. (5)). On tree-based graphs, this echo cancellation is required for correctness of exact probabilistic inference. In loopy graphs (without well-justified semantics), this term still compensates for the message a node $t$ would otherwise send to itself via a neighbor $s$ (thus the term "echo"). Experiments suggest that sometimes this term improves the label predictions, at other times not. We thus propose to use the EC term as an optional parameter to tweak existing algorithms.

---
[7]Notice that one could thus further generalize clamping by considering a vector $\boldsymbol{\gamma}$ with $0 \leq \gamma_i \leq 1$ that represents the strength of clamping for each node $i$. Then define $\bar{\mathbf{C}}^{(\boldsymbol{\gamma})} \triangleq \mathbf{I} - \text{diag}(\boldsymbol{\gamma})\mathbf{C}$. We do not investigate here this generalization any further.

$$E(\mathbf{F}) = ||\mathbf{F} - \mathbf{X} - \mathbf{W}^{(\alpha,\beta,\gamma)}\mathbf{FH}\underbrace{+\mathbf{D}^*\mathbf{FH}^2}_{\text{EC}}||^2 \quad (16)$$

Here $\mathbf{D}^*$ hides a number of details and is appropriately defined as $\mathbf{D}^* \triangleq \text{diag}(D_{11}^*, \ldots, D_{nn}^*)$ with $D_{ii}^* \triangleq \sum_j W_{ij}^{(\alpha,\beta,\gamma)} W_{ji}^{(\alpha,\beta,\gamma)}$.

Notice that this allows modifying existing algorithms in the obvious way. For example, by adding the echo cancellation term to the standard PageRank equation eq. (17), one can diminish the mutual PageRank increase from two directly connecting nodes:

$$\mathbf{f} \leftarrow \frac{\bar{\alpha}}{n}\mathbf{1} + \alpha \mathbf{W}^{\text{row}}\mathbf{f}\underbrace{-\alpha^2 \mathbf{D}^*\mathbf{f}}_{\text{EC}} \quad (17)$$

**Idea 6: Scaling modulation.** Our sixth idea is to introduce the *scaling factor* $\varepsilon$ into the update equations. As our experiments in section 6 show, this parameter allows us to increase the accuracy of label prediction with only few iterations and at one very well-defined value: the boundary of convergence and divergence of the update equations. Our intuition for this empirical observation is this boundary condition is an optimal tradeoff between two competing influences: (*i*) higher values allow information to propagate further; and (*ii*) lower values prevent creating adverse coupling effects. At the boundary, the label information can propagate far without yet creating adverse coupling effects. We call the multiplicative factor which exactly separates convergence from divergence, the "*convergence boundary*" $\varepsilon_*$. Choosing any $\varepsilon$ with $s \triangleq \frac{\varepsilon}{\varepsilon_*}$ and $s < 1$ guarantees convergence. We call any choice of $s$ the "*convergence parameter*".

**SSLH.** By putting these 6 ideas together, we now define *Semi-Supervised Learning with Heterophily (SSL-H)* as follows:

DEFINITION 3 (SSL WITH HETEROPHILY (SSL-H)). *Using above introduced notations,* Semi-Supervised Learning with Heterophily *is a family of label update equations given by:*

$$\mathbf{F} \leftarrow \mathbf{X} + \varepsilon \mathbf{W}^{(\alpha,\beta,\gamma)}\mathbf{FH}\underbrace{-\varepsilon^2 \mathbf{D}^*\mathbf{FH}^2}_{\text{EC}} \quad (18)$$

We also give sufficient and necessary convergence criteria, a closed-form solution, and the energy function that these update equations are minimizing.

THEOREM 4 (SSL WITH HETEROPHILY (SSL-H)).
*1) Convergence: The update equations converge if and only if*

$$\rho\left(\varepsilon \mathbf{H} \otimes \mathbf{W}^{(\alpha,\beta,\gamma)\mathsf{T}}\underbrace{+\varepsilon^2 \mathbf{H}^2 \otimes \mathbf{D}^*}_{\text{EC}}\right) < 1 \quad (19)$$

*2) Closed-form: If they converge, then they converge towards the closed-form solution given by:*

$$\text{vec}(\mathbf{F}) = (\mathbf{I} - \varepsilon \mathbf{H} \otimes \mathbf{W}^{(\alpha,\beta,\gamma)\mathsf{T}}\underbrace{+\varepsilon^2 \mathbf{H}^2 \otimes \mathbf{D}^*}_{\text{EC}})^{-1}\text{vec}(\mathbf{X}) \quad (20)$$

*3) Energy function: The closed-form minimizes the following energy function:*

$$E(\mathbf{F}) = ||\mathbf{F} - \mathbf{X} - \varepsilon \mathbf{W}^{(\alpha,\beta,\gamma)}\mathbf{FH}\underbrace{+\varepsilon^2 \mathbf{D}^*\mathbf{FH}^2}_{\text{EC}}||^2 \quad (21)$$

Figure 6 shows how existing SSL methods can be seen as special instances of SSL-H. Thus all these methods can be motivated from first principles as *approximations of standard BP on particular choices of graphs*. This connection was previously not known.

|  | clamping matrix $\mathbf{C}$ | propagation matrix $\mathbf{W}^*$ | compatibility matrix $\mathbf{H}$ | EC | energy function |
|---|---|---|---|---|---|
| HF [44] | hard | $\mathbf{W}^{\text{row}}$ | $\mathbf{I}$ | n | $\sum(\cdot)^2$ |
| LNP [38] | soft | $\mathbf{W}^{\text{row}}$ | $\mathbf{I}$ | no | $\approx \sum(\cdot)^2$ |
| LGC [42] | soft | $\mathbf{W}^{\text{red}}$ | $\mathbf{I}$ | no | $\sum(\cdot)^2$ |
| MRW [16] | soft | $\mathbf{W}^{\text{col}}$ | $\mathbf{I}$ | no | – |
| RNC [19] | soft | $\mathbf{W}$ | $\mathbf{I}$ | no | – |
| LinBP [9] | soft | $\mathbf{W}$ | $\mathbf{H}$ | yes/no | $||\cdot||^2$ |
| SSL-H | $\bar{\mathbf{C}}^{(\gamma)}$ | $\mathbf{W}^{(\alpha,\beta)}$ | $\mathbf{H}$ | yes/no | $||\cdot||^2$ |

Figure 6: Existing multi-class SSL methods are particular choices in a larger design space for SSL-H methods.

Notice that we took one liberty in interpretation with this table: The original formalization of RNC [18, 19] has a probabilistic interpretation of the label distribution and thus required normalization of the label distributions after each update. However, this normalization prevents the formalism from having convergence guarantees nor a closed-form solution (compare to BP that does not have a closed-form solution nor convergence guarantees either). We thus suggest to use residual labels in $\mathbf{X}$, to remove the normalizing step, and to interpret the resulting label distribution in the framework of SSL-H. Then RNC just becomes a special case of LinBP with the heterophily matrix being the identity matrix.

From eq. (18), we can now easily generalize the harmonic functions (HF) method [44] to "*harmonic functions with heterophily*" (HF-H) by choosing $\mathbf{H} \neq \mathbf{I}_k$. We can also create entirely novel variants; e.g., a soft-clamped variant for HF-H (by choosing $\gamma = 0$) or a hard-clamped variant of LinBP (by choosing $\gamma = 1$).

Finally notice that the propagation matrix is transposed in the closed-form (as indicated by the transpose symbol "$\mathsf{T}$.") This is not necessary for LinBP, LGC and any other variant in which the propagation matrix is symmetric.

## 5. COMPATIBILITY ESTIMATION

In this section, we focus on learning compatibility from partially labeled data *in a scalable way*. The model we like to learn are the independent parameters for the compatibility matrix $\mathbf{H}$. We refer to $\mathbf{H}$ as "heterophily matrix" to remind us the generalization that it allows. We start from the our previously derived energy minimization objective of SSL-H, derive our first method (LHE in section 5.1), and then improve on this method with in terms of speed (MHE in section 5.2) and accuracy (DHE in section 5.3).

### 5.1 Linear Heterophily Estimation (LHE)

We next derive "*Linear Heterophily Estimation*" (LHE) from our previously derived energy minimization objective of SSL-H. We start by drawing similarities between the energy functions for LLE (eq. (12)), LNP (fig. 4), and LinBP without EC (eq. (14)):[8]

$$E(\mathbf{W}) = \sum_{i=1}^n ||\mathbf{X}_{i:} - \sum_{j=1}^n W_{ij}^{\text{row}}\mathbf{X}_{j:}||^2 \quad \text{(LLE)} \quad (22)$$

$$E(\mathbf{F}) \approx ||\mathbf{F} - \mathbf{X}||^2 + \frac{\mu}{2}\sum_{i,j=1}^n W_{ij}||\mathbf{F}_{i:} - \mathbf{F}_{j:}||^2 \quad \text{(LNP)}$$

$$E(\mathbf{F}) = ||\mathbf{F} - \mathbf{X} - \varepsilon \mathbf{W}^{(\alpha,\beta,\gamma)}\mathbf{FH}||^2 \quad \text{(SSL-H)} \quad (23)$$

Notice that LLE tries to learn $\mathbf{W}$, whereas SSL-H and LNP try to learn $\mathbf{F}$. Also notice that by: (*i*) ignoring the explicit labels $\mathbf{X}$ (i.e., ignoring the fit term in LNP and ignoring the vector in SSL-H); (*ii*) the scaling factor $\varepsilon$; (*iii*) and the type of propagation matrix; and (*iv*) by replacing **H** with the identity matrix **I**), the right side of eq. (22) and eq. (23) become the same. Based on this comparison, we propose to learn the heterophily matrix **H** by minimizing the following energy function, subject to ($\mathbf{H} = \mathbf{H}^\mathsf{T}$) and ($\mathbf{H1} = \mathbf{1}$):

$$E(\mathbf{H}) = ||\mathbf{X} - \mathbf{W}^{(\alpha,\beta,\gamma)}\mathbf{X}\mathbf{H}||^2 \qquad \text{(LHE)} \qquad (24)$$

---

[8]In the remainder, we will ignore the optional EC term. In our experiments, we have not identified any parameter regime where including the EC term for propagation *consistently* outperforms alternatives without EC. However, the idea behind EC does become crucial for making DHE work in section 5.4 for learning.

This is justified for several reasons: (*i*) The difference between $E(\mathbf{X})$ and $E(\mathbf{F})$ on the left sides is not relevant for learning **H**. These are two different loss functions and we can focus only on one of them; (*ii*) $\varepsilon$ is a redundant parameter in this setting; (*iii*) We can use the mathematical machinery developed for LLE for our own problem setup; and (*iv*) This formalism explains heterophily for both LinBP and our generalization of SSL-H. At the same time, this formalism assures that the problem is convex with one well-specified optimum, and thus any standard optimizer can solve it.

**Unconstrained optimization.** We can easily transform the constraint optimization problem into an unconstrained one by posing the optimization problem simply over the $k^* \triangleq \frac{k(k-1)}{2}$ degrees of freedom of a $k \times k$-dimensional doubly stochastic matrix. Thus, let $\mathbf{h} \in \mathbb{R}^{k^*}$ and define **H** as function of the $k^*$ entries of **h**. The remaining matrix entries can be calculated in a straight-forward way from symmetry and normalization conditions. For example, for a $k = 3$, we have $k^* = 3$ degrees of freedom. Thus, **H** can be reconstructed from a 3-dimensional vector $\mathbf{h} = [h_1, h_2, h_3]^\mathsf{T}$ as follows:

$$\mathbf{H}(\mathbf{h}) = \begin{bmatrix} h_1 & h_2 & 1-h_1-h_2 \\ h_2 & h_3 & 1-h_2-h_3 \\ 1-h_1-h_2 & 1-h_2-h_3 & h_1+2h_2+h_3-1 \end{bmatrix}$$

We now have an unconstrained convex optimization problem:

$$E(\mathbf{h}) = ||\mathbf{X} - \mathbf{W}^{(\alpha,\beta,\gamma)}\mathbf{X}\mathbf{H}(\mathbf{h})||^2 \qquad \text{(LHE)} \qquad (25)$$

In our experiments with standard Python optimization libraries, we found that the latter formulation to be faster in practice.

## 5.2 Myopic Heterophily Estimation (MHE)

We next introduce a considerably faster method that we call "*Myopic Heterophily Estimation*" (MHE). We call it "myopic" as it tries to infer the relative frequencies between classes in the network by a straightforward frequency calculation between neighbors, followed by a transformation into a symmetric, doubly-stochastic matrix. The key motivation comes by observing that the formulation of eq. (24) requires that an iterative gradient descent algorithm evaluates the multiplication with the adjacency matrix $W$ in each iteration. Our goal is to derive an objective that factors out this calculation into one calculation before the actual iteration.

Given a partially labeled $n \times k$-matrix **X** with $X_{ic} = 1$ if node $i$ has label $c$ (recall that some nodes have no label, and thus their corresponding row $\mathbf{X}_{i:}$ is the null vector). Then the $n \times k$-matrix $\mathbf{N} \triangleq \mathbf{WX}$ has entries $N_{ic}$ representing the *number of labeled neighbors of node $i$ with label $c$*. Furthermore, the $k \times k$-matrix $\mathbf{M} \triangleq \mathbf{X}^\mathsf{T}\mathbf{N} = \mathbf{X}^\mathsf{T}\mathbf{WX}$ has entries $M_{cd}$ representing the *number of nodes with label $c$ that are neighbors of nodes with label $d$*. This symmetric matrix represents the observed frequencies of neighbors among the labeled nodes. Intuitively, we are trying to find a compatibility matrix which is "similar" to **M** in some way. We consider three intuitive variants of defining similarity with respect to **M**:

1. We make **M** row-stochastic by dividing each row by the sum of each row. The vector of row-sums can be expressed in matrix notation as **M1**. Thus define the "*observed compatibility matrix*":

$$\tilde{\mathbf{H}} = \text{diag}(\mathbf{M1})^{-1}\mathbf{M} \qquad \text{(Variant 1)} \qquad (26)$$

This matrix is row-stochastic (for each class $c$, the entry $\tilde{H}_{cd}$ gives the relative frequency of a it being connected to class $d$), but not doubly-stochastic. We propose to find the symmetric, doubly stochastic matrix **H** (i.e., it fulfills the $k^* \triangleq \frac{k(k+1)}{2}$ conditions implied by $\mathbf{H} = \mathbf{H}^\mathsf{T}$ and $\mathbf{H1} = \mathbf{1}$) that is closest to the observed compatibility-matrix $\tilde{\mathbf{H}}$ under the Frobenius norm as distance metric

$$E(\mathbf{H}) = ||\mathbf{H} - \tilde{\mathbf{H}}||^2 \qquad \text{(MHE)} \qquad (27)$$

This convex optimization problem of finding a Frobenius-norm optimum doubly stochastic approximation to a given matrix can be solved efficiently, for example, with an algorithm proposed in [41].

2. The other two variants use the same optimization criterium, but first calculate different symmetric, but non-stochastic observed compatibility matrices. The second variant uses the same normalization method as LGC [42] from section 2:

$$\tilde{\mathbf{H}} = \text{diag}(\mathbf{M1})^{-\frac{1}{2}}\mathbf{M}\text{diag}(\mathbf{M1})^{-\frac{1}{2}} \qquad \text{(Variant 2)} \qquad (28)$$

3. The third variant divides **M** by a number, s.t. the average matrix entry is $\frac{1}{k}$. This divisor is just the sum of all entries (in vector notation written as $\mathbf{1}^\mathsf{T}\mathbf{M1}$) divided by $k$:

$$\tilde{\mathbf{H}} = k(\mathbf{1}^\mathsf{T}\mathbf{M1})^{-1}\mathbf{M} \qquad \text{(Variant 3)} \qquad (29)$$

Notice that all three variants above have a second, deeper justification: On a *fully labeled graph*, each variant will learn the "correct" matrix, i.e. the matrix that best encodes the couplings between neighbors in a graph (we will verify this in our experiments where MHE is able to recover the compatibility matrix that was used for generating the fully labeled graph). However, MHE only considers the subgraph induced by the labeled nodes and thus requires a sufficient number of neighbors that are both labeled.[9]

## 5.3 Distant Heterophily Estimation (DHE)

We will introduce a method called "Distant Heterophily Estimation" (DHE) which uses some kind of "*smoothing*" that helps to take into account more than just the nearest neighbor. In a graph with $m$ edges and a small fraction $f$ of labeled nodes, the number of labeled neighbors may be quite small too ($\sim mf^2$). Yet the number of "distance2-neighbors" (i.e. nodes which are connected via a path of length 2) is higher in proportion to the average node degree $d$ ($\sim dmf^2$). Similarly for distance-$\ell$-neighbors ($\sim d^{\ell-1}mf^2$). Can we leverage this information to improve our estimate of **H**?

Notice that information that travels via a path of length $\ell$ gets modulated $\ell$ times, i.e. via a power of the modulation matrix: $\mathbf{H}^\ell$. We propose to use powers of the matrix **H** to be estimated and compare them against an observed "*length-$\ell$ compatibility matrix*."

Powers of the adjacency matrix $\mathbf{W}^\ell$ with entries $W_{ij}^\ell$ describes the number of paths of length $\ell$ between any nodes $i$ and $j$. Similarly to before, let $\mathbf{N}^{(\ell)} \triangleq \mathbf{W}^\ell\mathbf{X}$ and $\mathbf{M}^{(\ell)} \triangleq \mathbf{X}^\mathsf{T}\mathbf{N}^{(\ell)} = \mathbf{X}^\mathsf{T}\mathbf{W}^\ell\mathbf{X}$ with entries $M^{(\ell)}{}_{cd}$ representing the number of nodes of class $d$ that are neighbors of nodes of class $c$. Normalize this matrix in any of the previous three variants to $\tilde{\mathbf{H}}^{(\ell)}$, and then minimize following "*distance-smoothed*" energy function

$$E(\mathbf{H}) = \sum_{\ell=1}^{\ell_{\max}} w_\ell ||\mathbf{H}^\ell - \tilde{\mathbf{H}}^{(\ell)}||^2 \qquad (30)$$

---
[9]Recall from the comparison between inductive and transductive inference in the introduction that much of the predictive power in SSL methods comes from the availability of a large number of unlabeled data. MHE entirely ignores the graph structure given by unlabeled data.

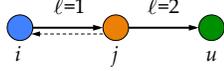

Figure 7: Illustration for non-backtracking paths

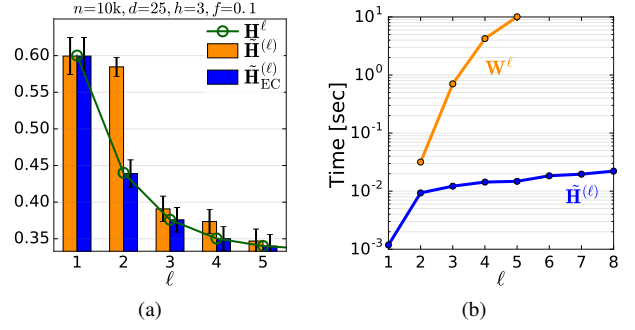

Figure 8: (a): Example 6: Using non-backtracking paths $\tilde{\mathbf{H}}_{\text{EC}}^{(\ell)}$ leads to an unbiased estimator, in contrast to using all paths $\tilde{\mathbf{H}}^{(\ell)}$. (b): Example 7: Calculating paths of increasing length $\ell$ quickly becomes infeasible since $\mathbf{W}^\ell$ becomes less sparse with $\ell$. In contrast, our proposed sequence for calculating $\tilde{\mathbf{H}}^{(\ell)}$ avoids calculating $\mathbf{W}^\ell$ explicitly and is super fast.

where the vector $\mathbf{W}$ of weights is a trade-off between having more data points for higher distances $\ell$ in proportion to $d$, but an attenuated signal. To parameterize the weight vector, we will use a "*scaling factor*" $\lambda$ defined by $w_{\ell+1} = \lambda w_\ell$. For example, a distance-3 weight vector will then be $[1, \lambda, \lambda^2]^\top$. We will comment on the optimal choice of this scaling factor during the experiments.

## 5.4 Non-backtracking paths

In our previous approach of learning from more distant neighbors, we made a slight but consistent mistake. We illustrate this mistake with the help of fig. 7. Consider the blue node $i$ which has one orange neighbor $j$, which has two neighbors, one of which is green node $u$. Then the blue node $i$ has one distance-2 neighbor $u$ that is green. However, our previous approach will consider all length-2 paths, one of which will lead back to node $i$. Thus, the row entry for node $i$ in $\mathbf{N}$ is $\mathbf{N}_{i:}^{(2)} = [1, 0, 1]$ (assuming blue, orange, and green represent classes 1, 2, and 3, respectively). In other words, $\mathbf{M}^{(2)}$ will always overestimate the diagonal entries.

What we propose is to ignore echos and to only consider "non-backtracking paths" in the powers of the adjacency matrix, which we denote with a subscript EC (for "*echo cancellation*"). Thus, we replace $\mathbf{W}^\ell$ with $\mathbf{W}_{\text{EC}}^{(\ell)}$ and calculate $\tilde{\mathbf{H}}_{\text{EC}}^{(d)}$ from $\mathbf{M}_{\text{EC}}^{(\ell)} \triangleq \mathbf{X}^\top \mathbf{W}_{\text{EC}}^{(\ell)} \mathbf{X}$.

For example, $\mathbf{W}_{\text{EC}}^{(2)} = \mathbf{W}^2 - \mathbf{D}$ (a node $i$ with degree $d_i$ has $d_i$ as the diagonal entry $W_{ii}$ in $\mathbf{W}$). Length-3 non-backtracking paths are already more involved: $\mathbf{W}_{\text{EC}}^{(3)} = \mathbf{W}^3 - (\mathbf{DW} + \mathbf{WD} - \mathbf{W})$. Longer length paths can be calculated recursively as follows:

PROPOSITION 5 (NON-BACKTRACKING PATHS). *Let $\mathbf{W}_{\text{EC}}^{(\ell)}$ be the matrix with $W_{\text{EC}\,ij}^{(\ell)}$ being the number of non-backtracking paths of length $\ell$ from node $i$ to $j$. Then $\mathbf{W}_{\text{EC}}^{(\ell)}$ for $\ell \geq 3$ can be calculated from following recurrence relation:*

$$\mathbf{W}_{\text{EC}}^{(\ell)} = \mathbf{W}\mathbf{W}_{\text{EC}}^{(\ell-1)} - (\mathbf{D} - \mathbf{I})\mathbf{W}_{\text{EC}}^{(\ell-2)}$$

*with starting values $\mathbf{W}_{\text{EC}}^{(1)} = \mathbf{W}$ and $\mathbf{W}_{\text{EC}}^{(2)} = \mathbf{W}^2 - \mathbf{D}$.*

EXAMPLE 6 (NON-BACKTRACKING PATHS). *Consider $\mathbf{H} = \begin{bmatrix} 0.1 & 0.8 & 0.1 \\ 0.8 & 0.1 & 0.1 \\ 0.1 & 0.1 & 0.8 \end{bmatrix}$. Then $\mathbf{H}^2 = \begin{bmatrix} 0.66 & 0.17 & 0.17 \\ 0.17 & 0.66 & 0.17 \\ 0.17 & 0.17 & 0.66 \end{bmatrix}$, and the top entry of the first row (permuting between first and second position) follows the series $0.8, 0.66, 0.562, 0.4934, \ldots$ for increasing exponent $\ell$ (shown as continuous line $\mathbf{H}^\ell$ in fig. 8a). We created synthetic graphs with $n = 10000$ nodes, average node degree $d = 10$ and powerlaw degree distribution with coefficient $-0.3$, compatibility matrix $\mathbf{H}$ with $h = 3$ (see section 6 for details on the synthetic graph generator), and removed the labels from $1 - f = 90\%$ of nodes Based on 1000 repetitions, we calculated the mean of the previously mentioned entry in both $\tilde{\mathbf{H}}^{(\ell)}$ and $\tilde{\mathbf{H}}_{EC}^{(\ell)}$. The the two bars in fig. 8a show the corresponding matrix entries, illustrating that the approach based on non-backtracking paths leads to a consistent estimate, in contrast to the full paths.*

**Efficient calculation of $\tilde{\mathbf{H}}_{\text{EC}}^{(\ell)}$.** Calculating either $\tilde{\mathbf{H}}^{(\ell)}$ or $\tilde{\mathbf{H}}_{\text{EC}}^{(\ell)}$ requires multiplications of multiple matrices. While matrix multiplication is associative, and the order in which we perform the multiplications does not change the result, it does considerably determine the time to evaluate a product. A naive evaluation strategy can quickly become problematic for increasing $\ell$.

We illustrate with $\mathbf{M}^{(4)}$: a default strategy is to first calculate $\mathbf{W}^{(4)} = \mathbf{W}(\mathbf{W}(\mathbf{W}\mathbf{W}))$ and then $\mathbf{M}^{(4)} = \mathbf{X}^\top(\mathbf{W}^{(4)}\mathbf{X})$. The problem is that the intermediate result $\mathbf{W}^{(\ell)}$ can be exponentially large. Concretely, if $\mathbf{W}$ is sparse with $m$ entries and average node degree $d$, then $\mathbf{W}^2$ has in the order of $d$ more entries entries ($\sim dm$), and $\mathbf{W}^\ell$ exponential more entries ($\sim d^{\ell-1}m$).

Thus intuitively, we like to choose the evaluation order so that intermediate results are as sparse as possible.[10] In our case, the ideal way to calculate the expressions is to keep $n \times k$ intermediate matrices as in $\mathbf{M}^{(4)} = \mathbf{X}^\top(\mathbf{W}(\mathbf{W}(\mathbf{W}(\mathbf{W}\mathbf{X}))))$. The way we propose to calculate $\tilde{\mathbf{H}}_{\text{EC}}^{(\ell)}$ is to first iteratively calculate $\tilde{\mathbf{N}}_{\text{EC}}^{(\ell)} = \mathbf{W}\tilde{\mathbf{N}}_{\text{EC}}^{(\ell-1)} - (\mathbf{D} - \mathbf{I})\tilde{\mathbf{N}}_{\text{EC}}^{(\ell-2)}$ with starting values $\tilde{\mathbf{N}}_{\text{EC}}^{(1)} = \mathbf{W}\mathbf{X}$ and $\tilde{\mathbf{N}}_{\text{EC}}^{(2)} = \mathbf{W}\tilde{\mathbf{N}}_{\text{EC}}^{(1)} - \mathbf{D}\mathbf{X}$, then $\tilde{\mathbf{M}}_{\text{EC}}^{(\ell)} = \mathbf{X}^\top\tilde{\mathbf{N}}_{\text{EC}}^{(\ell)}$, and finally $\tilde{\mathbf{H}}_{\text{EC}}^{(\ell)}$ from normalizing $\tilde{\mathbf{M}}_{\text{EC}}^{(\ell)}$.

EXAMPLE 7 (EFFICIENT CALCULATION OF $\tilde{\mathbf{H}}_{\text{EC}}^{(\ell)}$). *Using the parameters from example 6, fig. 8b shows the times it takes to evaluate $\mathbf{W}^\ell$ against the more efficient evaluation strategy for $\tilde{\mathbf{H}}_{EC}^{(\ell)}$. Notice that calculating $\mathbf{W}^\ell$ for such graphs with powerlaw degree distribution scales worse than for uniform degree distribution.*

## 6. EXPERIMENTS

**Questions.** Our experiments focus on following three questions: (1) How well do the various propagation schemes propagate a known heterophily as function of relevant problem parameters? (2) How well do the various estimation methods estimate heterophily in a partially labeled graph as function of problem parameters, and (3) How scalable are the methods?

**Experimental protocol.** We chose to evaluate our techniques in a completely controlled simulation environment, as this allows us to change parameters of the ground truth and evaluate the accuracy of our techniques as result of systematic changes to such parameters. In other words, this allows us to explore the methods

---

[10] This observation is well known in linear algebra and is analogous to query optimization in relational algebra: The two query plans $R(x) \bowtie (\pi_x S(x,y))$ and $\pi_x(R(x) \bowtie S(x,y))$ return the same values, but the former can be considerably slower. Similarly, the "evaluation plans" $(\mathbf{W}\mathbf{W})\mathbf{X}$ and $\mathbf{W}(\mathbf{W}\mathbf{X})$ are algebraically equivalent, but the former is considerably slower for sparse and $(n \times n)$ $\mathbf{W}$, and $(n \times k)$ $\mathbf{X}$ with $n \gg k$.

| | | | |
|---|---|---|---|
| **Problem parameters** | $n$ | number of nodes | $10^4, 10^2 - 10^6$ |
| | $d$ | average degree | 5, 10, 25 |
| | dist | degree distribution | powerlaw with exponent -0.3 |
| | $h$ | strength of compatibility | 2, 5, 8 |
| | $f$ | fraction of labels | $0.05, 0.001 - 0.9$ |
| **Solution parameters** | $\alpha, \beta$ | propagation matrix | 0, 0.5, 1 |
| | $\gamma$ | clamping | 0, 0.5, 1 |
| | $s$ | convergence parameter | $0.1 - 10$ |
| | $r$ | number of iterations | $1 - 20$ |
| | EC | echo cancellation | yes / no |

Figure 9: Key parameters and typical values used in the experiments.

in a well-controlled, reproducible setting and make general observations that would not be apparent when examining only particular individual data set on their own. Real networks usually have a powerlaw degree distribution with many low-degree vertices. In contrast, the well-studied stochastic block-model [1] leads to networks with degree distributions that are not similar to those found in most empirical network data. Our synthetic graph generator is thus a variant thereof with two important differences: (1) We *actively control the degree distributions* in the resulting graph; and (2) we *plant exact graph properties* (instead of fixing a property only in expectation). In other words, our generator preserves natural desired degree distribution and "plants" a compatibility matrix during graph generation, which allows us to control all important problem parameters. The input to the generator is a tuple of parameters $(n, m, \alpha, \mathbf{H}, \text{dist})$ where $n$ is the number of nodes, $m$ is the number of edges, $\alpha$ is the node label distribution with $\alpha(i)$ being the fraction of nodes of class $i$ ($i \in [k]$), $\mathbf{H}$ is a doubly stochastic symmetric compatibility matrix, edge potential, and dist is a family of degree distributions (e.g., uniform or power law with specified coefficient). Details of the data generator are found in the online appendix

**Parameter choices and quality assessment.** Throughout our experiments, we use $k = 3$ classes and the following compatibility matrix $\mathbf{H} = \begin{bmatrix} 1 & h & 1 \\ h & 1 & 1 \\ 1 & 1 & h \end{bmatrix}$, parameterized by a value $h$ representing the ratio between min and max entries. Thus parameter $h$ models the *strength* of the potential. For example, $\mathbf{H} = \begin{bmatrix} 0.1 & 0.8 & 0.1 \\ 0.8 & 0.1 & 0.1 \\ 0.1 & 0.1 & 0.8 \end{bmatrix}$ for $h = 8$, and $\mathbf{H} = \begin{bmatrix} 0.2 & 0.6 & 0.2 \\ 0.6 & 0.2 & 0.2 \\ 0.2 & 0.2 & 0.6 \end{bmatrix}$ for $h = 3$. We create graphs with $n$ nodes and assign each equal fractions of nodes to one of the 3 classes: $\alpha = [\frac{1}{3}, \frac{1}{3}, \frac{1}{3}]$. We also vary the parameter $m$ and use $d = 2\frac{m}{n}$ as average degree in the graph as parameter for the experiments and assume a uniform distribution or a power law distribution with coefficient 0.3. We then remove a number of nodes so that only a fraction $f$ of the nodes have labels and test accuracy of a method based on the hold-out set. For each node in the hold out set, we calculate the label with maximum belief as predicted by a method and then evaluate labeling "accuracy" as the percentage of correctly retrieved labels on the hold-out set only. Notice that accuracy of $\frac{1}{3}$ corresponds to random assignment of labels. Key parameters and their values are reported in fig. 9.

**Computational setup and code.** We implemented our algorithms in Python that uses optimized libraries for sparse matrix operations.[11] The implementation runs on a 2.5 Ghz Intel Core i5 with 16G of main memory and a 1TB SSD hard drive. To allow comparability across implementations, we limit evaluation to one processor. We use fmin_slsqp in scipy.optimize to minimize our function using Sequential Least SQuares Programming. For calculating the approximate spectral radius of a given matrix, we use a fast approximate method from the PyAMG library [3] that implements a technique described in [2]. Our code including the data generator is inspired by Scikit-learn [28] and will be made available to encourage reproducible research.

### 6.1 Accuracy for heterophily propagation

Each subfigure of row 1 in fig. 10 shows the average accuracy as function of the numer of iterations $r$ (between 1 and 20) and the convergence parameter $s$ (between 0.1 and 10). The figures also display the unmodified convergence parameter $s_0$, the number of simulations for each data point, and the standard deviation (std) for accuracy for one choice of $r$ ($r = 3$ for $n = 10^4$ and $r = 3$ for $n = 10^3$). The chosen parameters are $n = 10^4, h = 5, d = 25, f = 0.05$. Our experiments lead to following insights:

**1. For strong potentials ($h \uparrow$), few iterations ($r \downarrow$) and a higher convergence parameter ($s > 1$) can give higher accuracy than iterating until convergence with $s = 0.5$.** Figure 10a shows an instance where iterating only 4 times and using $s = 3$ instead of $s = 0.5$ increases the accuracy from 0.85 to 0.969. The conclusion is that we do not need to run until convergence for maximal accuracy.

**2. For strong potentials ($h \uparrow$), normalized propagation matrices ($\alpha + \beta > 0$) and clamping ($\gamma > 0$) can increase accuracy.** Rows 1 in fig. 10 shows examples with $h = 5$ where normalized propagation matrices ($\mathbf{W}^{\text{col}}$ and $\mathbf{W}^{\text{red}}$) and clamping ($\gamma = 1$) leads to higher accuracy. The intuition is that with $s_0 = 17.16$, the unnormalized propagation matrix would have to be weakened by a factor $> 17$ for the update equations to converge. By avoiding to iterate until convergence, we can choose a higher convergence parameter. In addition, clamping also reduces the effect of cycles ($s_0 = 16.34$). Notice that the accuracy increases from 0.85 to 0.99.

**3. For weak potentials ($h \downarrow$) or high fraction of labeled nodes ($f \uparrow$), normalized propagation matrices and clamping do not necessarily work better.** Not shown is that for weak potentials ($h = 2$), $\mathbf{W}$ works similarly well as the normalized variants, and it appears that $s = 0.5$ as suggested in [9] work rather robustly.[12] Also not shown is that for high fraction of labeled nodes ($f = 0.9$), clamping can reduce the accuracy and $\mathbf{W}$ works better than $\mathbf{W}^{\text{red}}$.

**4. There is no strong empirical support for the use of the echo cancellation term.** Overall we had difficult time finding any parameter choice for which the echo cancellation term can add additional benefit to other recommendations.[13] For this reason, we recommend to ignore the echo cancellation term.

### 6.2 Results on heterophily estimation

**5. DHE works best for variant 1 and increasing $\ell_{\max}$ and scaling factor in the order of $\lambda = 10$.** In fig. 10e, DHE is used with our three variants and different maximal path lengths $\ell_{\max}$ to estimate the compatibility matrix. The vertical axis shows the $L^2$-norm between estimation and GT. We observed that variant 3 generally performs worse, and variant 2 generally has higher variance, which is why we propose variant 1. We also see that increasing $\ell_{\max}$ to 5 generally works best. However $\ell_{\max}$ above a certain threshold around 100 becomes unstable (fig. 10f). This observation holds over a wide range of $d$ and $f$ (fig. 10g). Figure 10h shows the advantage of using $\ell_{\max} = 5, \lambda = 10$ for estimating $\mathbf{H}$ as compared to just MHE, i.e. DHE with $\ell_{\max} = 1$.

**6. DHE works well enough to allow estimation and labeling on the same graph with high accuracy.** Figure 10i and fig. 10j show results from first estimating $\mathbf{H}$ from a partially labeled graph, and then labeling the remaining nodes with SSLH and parameters $(\alpha, \beta, \gamma) = (0, 1, 0.5)$ with $s = 3$ for $h = 3$ and $s = 5$ for $h = 8$ and 4

---
[11]Scipy and Numpy: http://docs.scipy.org/

[12]On a side note, notice that $\mathbf{W}^{\text{red}}$ converges after only $r = 3$ iterations in row 2, whereas $\mathbf{W}$ needs $r = 4$ iterations for $s = 0.5$.
[13]If we fix $s = 0.5$, then the EC term can sometimes improve the accuracy.

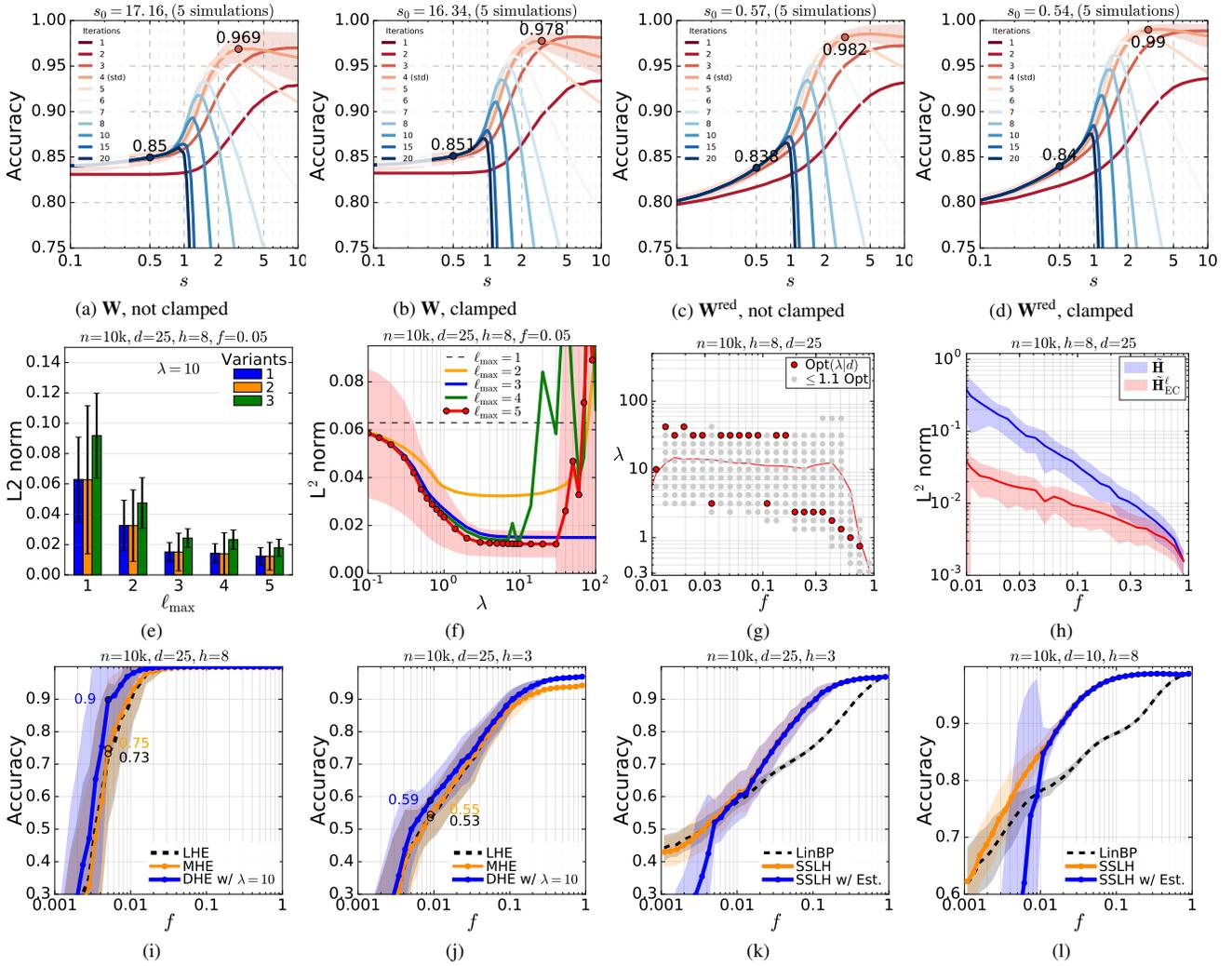

Figure 10: (a)-(d): Section 6.1: Using a convergence parameter that would not make the iterations converge ($s > 1$) with few iterations ($r = 4$), together different propagation matrices ($\alpha$ or $\beta \neq 0$), and clamping ($\gamma \neq 0$) increased accuracy from 0.85 to 0.99 in the case of a strong coupling ($h = 5$) and dense network ($d = 25$). (e)-(l): Section 6.2: DHE works generally better with increasing maximal path length $\ell_{max}$ (a) and scaling factor $\lambda$ in the order of 10 (b). This holds for a broad range or parameters, such as $f$ (c) and allows the compatibility matrix to be estimated with higher accuracy, especially for small $f$ (d). As result, the predicated labels by using DHE can better (e, f), and lead to the intersting result that using DHL to estimate the compatibility matrix and then SSLH to label the remaining nodes can be *better* than using the existing method LinBP without knowing the compatibility matrix.

iterations. We see that DHE can outperform MHE and LHE, which both work comparably.

**7. Combining DHE with SSLH on a graph with unknown heterophily H can work better than LinBP and knowing H.** Figures 3a, 10k and 10l from the introduction show that DHE combined with SSLH can solve the problem of labeling the remaining nodes with an accuracy higher than LinBP (i.e. SSLH with $(\alpha, \beta, \gamma) = (0,0,0)$ and $s = 0.5$ and 10 iterations) even when the compatibility matrix is not known to SSLH, but to LinBP.

## 6.3 Results on Scalability

**8. We can estimate H in a fraction that it takes to label the remaining nodes.** Our overview fig. 3b from the introduction shows the enormous scalability of our combined methods. On our largest graph with 400k nodes and 5M edges, propagation with 10 iterations took 56 sec, estimation of the ideal $\varepsilon_{max}$ 39 sec, estimation of the compatibility matrix with LHE 14 sec, MHE only 0.4 sec, and our suggested method DHE with best accuracy and still superior speed 2 sec. Notice that this last result that learning heterophily from a partially labeled graph is more than one order faster time than later propagating it is one of our key surprising results.

## 7. CONCLUSIONS

This paper proposes a novel SSL formulation that allows us to use not only similarity and dissimilarity, but any type of mutual coupling strengths between different classes of nodes (we abstract this with the doubly stochastic compatibility matrix). We show how our formulation generalizes a number of existing frameworks and naturally extends them from the commonly used "smoothness assumption" to heterophily. Notably, this approach includes a recent approach called Linearized Belief Propagation as special case and improves on its prediction accuracy. The paper also showed how to estimate the compatibility matrix based on partially labeled data in a fraction of the time it takes to later label the remaining data. We intend to polish and release our code as part of the open-source Scikit-learn Python library [31] to allow reproducible results.

**Acknowledgements.** This work was supported in part by NSF grant IIS-1553547. I would like to thank Christos Faloutsos for continued support and Stephan Günneman for insightful comments on an early version of this paper.

# APPENDIX

## A. USEFUL MODULATION PROPERTIES OF VECTORS AND MATRICES

Multiplication of a vector with a matrix corresponds to a linear mapping (also called linear transformation) of the vector. Such linear transformations preserve the operations of vector addition and scalar multiplication [48]. In the following, we will expose important "modulation properties" of certain matrices that are at the core of semi-supervised learning with heterophily.

The first fact is widely used for the theory of Markov chains. We state the theory in a bit more general way.

LEMMA 8 (MODULATION BY A ROW-SUM-CONSTANT MATRIX). *Linearly mapping a row-vector with row-sum $r$ with a matrix with constant row-sum $r$ leads to a vector with row-sum $rs$.*

PROOF. Let $\mathbf{x} = [x_1, \ldots, x_k]$ be a $k$-dimensional row vector whose entries sum to $s$ ($\sum_{i=1}^{k} x_i = s$), and $\mathbf{H}$ to be a $[k \times k]$ matrix with constant row-sum $r$ ($\sum_{i=1}^{k} H_{ji} = r$). Consider the $k$-dimensional row vector $\mathbf{f}$ resulting from "modulating" (linearly transforming) $\mathbf{x}$ by $\mathbf{H}$. We show that the entries of $\mathbf{f}$ also sum to $rs$:

$$\mathbf{f} = \mathbf{x}\mathbf{H}$$
$$f_i = \sum_\ell x_\ell H_{\ell i}$$
$$\sum_i f_i = \sum_i \sum_\ell x_\ell H_{\ell i}$$
$$= \sum_\ell x_\ell \underbrace{\sum_i H_{\ell i}}_{=r} = r\sum_\ell x_\ell = rs$$

| | |
|---|---|
| $n, n_\ell$ | number of nodes, or labeled nodes |
| $i, j, v$ | indices used for nodes |
| $k$ | number of classes |
| $c$ | index used for classes |
| $\mathbf{I}$ | $n \times n$ diagonal identity matrix |
| $\mathbf{W}$ | $n \times n$ symmetric adjacency matrix (data graph) |
| $\mathbf{D}$ | $n \times n$ diagonal degree matrix |
| $\mathbf{L}$ | $n \times n$ Laplacian matrix: $\mathbf{D} - \mathbf{W}$ |
| $\mathbf{L}_n$ | $n \times n$ normalized Laplacian matrix: $\mathbf{I} - \mathbf{W}_{\text{red}}$ |
| $\mathbf{W}_{\text{row}}$ | $n \times n$ row stochastic adjacency matrix: $\mathbf{D}^{-1}\mathbf{W}$ |
| $\mathbf{W}_{\text{red}}$ | $n \times n$ row stochastic adjacency matrix: $\mathbf{D}^{-\frac{1}{2}}\mathbf{W}\mathbf{D}^{-\frac{1}{2}}$ |
| $\mathbf{S}$ | $n \times n$ diagonal seed label indicator matrix |
| $\mathbf{C}$ | $n \times n$ diagonal hard clamping matrix |
| $\mathbf{X}$ | $n \times k$ a priori label matrix |
| $\mathbf{F}$ | $n \times k$ inferred label matrix |
| $\mathbf{H}$ | $k \times k$ compatibility (heterophily, affinity, or coupling) matrix |
| $\varepsilon$ | Scaling factor for $\mathbf{H}$ |
| $\tilde{\mathbf{W}}^{(\ell)}$ | $n \times n$ non-backtracking walk matrix for path length $\ell$ |
| $\tilde{\mathbf{H}}_{\text{EC}}^{(\ell)}$ | $k \times k$ observed compatibility matrix with EC for path length $\ell$ |
| $\mathbf{X}^\mathsf{T}$ | transpose of matrix $\mathbf{X}$ |
| $\text{vec}(\mathbf{X})$ | vectorization of matrix $\mathbf{X}$ |
| $\mathbf{X} \otimes \mathbf{Y}$ | Kronecker product between matrices $\mathbf{X}$ and $\mathbf{Y}$ |
| $\rho(\mathbf{X})$ | spectral radius of a matrix $\mathbf{X}$ |
| $\mathbf{X}_{i:}, \mathbf{X}_{:j}$ | $i$-th row vector, and $j$-th column vector of $\mathbf{X}$ |
| $\|\mathbf{X}\|$ | Frobenius norm of matrix $\mathbf{X}$: $\sqrt{\sum_{i,j}|X_{ij}|^2}$ |

Table 1: Nomenclature

Notice that the same also applies to column-vectors modulated "from the left" by column-stochastic matrices. This follows immediately from the transpose: $\mathbf{f}^\mathsf{T} = \mathbf{H}^\mathsf{T}\mathbf{x}^\mathsf{T}$. □

The case for $r = s = 1$ covers the well-known from Markov chains: Mapping a stochastic row-vector with a row-stochastic matrix from the right leads to another row-stochastic vector. The additional condition that no entry of the resulting vector is negative follows from the non-negativity of the input and the positivity of multiplication.

The case for $r = s = 0$ has been implicitly used in recent work that proposed the linearization of Belief Propagation [9]: Mapping a residual row-vector with a residual matrix from the right leads to another residual vector.

The second fact has to do with residual vectors. The original LinBP paper [9] proposed to center both, beliefs $\mathbf{f}$ and the heterophily matrix $\mathbf{H}$. We next show that centering the heterophily matrix is actually not required. We again pose the underlying idea as slightly more general:

LEMMA 9 (MODULATING RESIDUAL VECTORS). *Consider a residual row vector modulated by an arbitrary matrix. Adding or subtracting a constant value to each entry of the matrix does not change the resulting modulated vector.*

PROOF. Let $\mathbf{x} = [x_1, \ldots, x_k]$ be a $k$-dimensional residual row vector ($\sum_{i=1}^k x_i = 0$), and $\mathbf{H}$ to be a $[k \times k]$ matrix. Consider the $k$-dimensional row vector $\mathbf{f}$ resulting from transforming $\mathbf{x}$ by $\mathbf{H}$, and the row vector $\mathbf{f}'$ resulting from transforming $\mathbf{x}$ by $\mathbf{H}'$ where $H'_{ij} = H_{ij} + \delta$. It is easy to show that that $\mathbf{f}' = \mathbf{f}$:

$$\mathbf{f}' = \mathbf{x}\mathbf{H}'$$
$$= \mathbf{x}(\mathbf{H} + \delta\mathbf{1}_{k\times k})$$
$$= \mathbf{f} + \delta\underbrace{\mathbf{x}\mathbf{1}_{k\times k}}_{=\mathbf{0}} = \mathbf{f}$$

Notice that $\mathbf{x}$ needs to be a residual vector (i.e. centered around 0) for this equivalence to hold. □

A consequence is that our formulation of SSL-H (and thus also the formulation LinBP) works identically if the heterophily matrix $\mathbf{H}$ is *not* centered but instead kept as a doubly-stochastic matrix. Thus, the centering of both beliefs $\mathbf{f}$ and the heterophily matrix $\mathbf{H}$ as suggested in [9] is actually not necessary. Instead, if the relative frequencies by which different node classes connect to each other is known, then this matrix can be used *without centering* during the linearized propagation and will lead to identical results and thus node labels.

EXAMPLE 10. *Consider the residual vector $\mathbf{x}_2 = [2, -1, -1]$ and the row-stochastic heterophily matrix $\mathbf{H}_2 = \begin{bmatrix} 0.1 & 0.8 & 0.1 \\ 0.8 & 0.1 & 0.1 \\ 0.1 & 0.1 & 0.8 \end{bmatrix}$ from fig. 2c. Then $\mathbf{f}_2 = \mathbf{x}_2\mathbf{H}_2 = [-0.7, 1.4, -0.7]$. Now consider the residual matrix $\mathbf{H}'$ after centering around $\frac{1}{3}$ (i.e. $H'_{ij} = H_{ij} - \frac{1}{3}$). Then $\mathbf{f}'_2 = \mathbf{x}_2\mathbf{H}'_2$ is still $[-0.7, 1.4, -0.7]$.*

Despite this equivalence, we still suggest to learn residual heterophily matrices. The reasons are: 1) the scaling of heterophily matrices described in eq. (11) and during our experiments are cleaner to formalize with centered matrices; and 2) entries in a residual matrix are not required to follow any domain conditions, in contrast to stochastic matrices for which the probabilistic interpretation requires all entries to be $\geq 0$. This lack of a non-negativity constraint allows learning to happen slightly faster, in practice.

## B. PROOF LINBP LOSS FUNCTION

PROOF THEOREM 1. It was shown in [9] that if the update equations eq. (8) converge, then they converge towards the solution to the following equation system that is equivalent to eq. (10):

$$\mathbf{F} = \mathbf{X} + \mathbf{W}\mathbf{F}\mathbf{H}\underbrace{-\mathbf{D}\mathbf{F}\mathbf{H}^2}_{\text{EC}} \quad (31)$$

Observe now that if eq. (31) holds, then the energy function eq. (14) becomes zero as $\mathbf{F} - \mathbf{X} - \mathbf{W}\mathbf{F}\mathbf{H}\underbrace{+\mathbf{D}\mathbf{F}\mathbf{H}^2}_{\text{EC}} = 0$. On the other hand, the energy function is quadratic and can never become negative, and its possible minimum is zero. And when it becomes zero, then also eq. (31) holds.

Thus, minimizing eq. (14) also leads to the solution after convergence of the update equations. □

## C. PROOF SSL WITH HETEROPHILY

PROOF THEOREM 4. By using the substitutions $\mathbf{W} \leftarrow (\bar{\mathbf{C}}\mathbf{W}^*)$ and $\mathbf{H} \leftarrow (\varepsilon\mathbf{H})$, claim 1 follows from eq. (9), claim 2 follows from eq. (10), and claim 3 follows from theorem 1. □

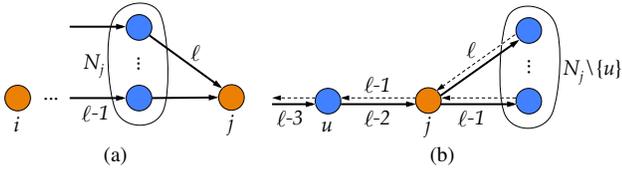

Figure 11: Illustrations for proof of proposition 5.

## D. PROOF NON-BACKTRACKING PATHS

PROOF PROPOSITION 5 (NON-BACKTRACKING PATHS). We prove by induction on $\ell$, the length of the paths. First notice that $\mathbf{W}_{\text{EC}}^{(1)} = \mathbf{W}$, because there is no back-tracking path of length 1. Also, $\mathbf{W}_{\text{EC}}^{(2)} = \mathbf{W}^2 - \mathbf{D}$, because $\mathbf{W}^2$ includes the only feasible backtracking paths of length 2 on its diagonal $W_{ii} = d_i$ where $d_i$ is the degree of node $i$.

Next assume the recurrence relations holds for paths of length up to $\ell - 1$. Then the number of non-backtracking paths of length $\ell$ that start at node $i$ and arrive at node $j$ consists of two parts:

(1) The number of paths that have not backtracked at previous steps and that arrive at node $j$ at step $\ell$ are $\sum_{u \in N_j} W_{\text{EC}_{iu}}^{\ell-1}$, where $N_j$ stands for the neighbors of $j$ (see fig. 11a).

(2) Among those, we need to subtract the number of walks that backtrack at the $\ell$-th step. Those paths need to pass through $j$ at step $\ell - 2$. Consider such a path that passes through a neighbor $u \in N_j$ at step $\ell - 3$ (see fig. 11b). At step $\ell - 1$ there are $d_j - 1$ paths that continue from node $j$ to neighbors (because the walk that would backtrack to $u$ is forbidden) and then come back at step $\ell$. Thus we need to subtract $W_{\text{EC}_{ij}}^{(\ell-2)}(d_j - 1)$ such paths. This gives us

$$W_{\text{EC}_{ij}}^{(\ell)} = \sum_{u \in N_j} W_{\text{EC}_{iu}}^{(\ell-1)} - W_{\text{EC}_{ij}}^{(\ell-2)}(d_j - 1)$$

which leads to our recurrence relation in matrix notation. □

## E. EXISTING GRAPH GENERATORS AND HARDNESS OF NODE LABELING

There is a large body of work that proposes various realistic synthetic generative graph models. However, almost all of this work assumes unlabeled graphs. While one could use these existing graph generators to have realistic graphs, one cannot easily take a graph and then label the nodes according to some desired compatibility matrix. In fact, this problem is NP-hard.

PROPOSITION 11 (LABELING WITH POTENTIALS). *Given a graph $G(V,E)$. Finding labels $\ell : v \in V \to [k]$ so that the labels follow a given stochastic affinity matrix $\mathbf{H}$ (where $H(i,j)$ determines the average fraction of nodes of class $j$ connected to a node of class $i$) is NP-hard.*

PROOF PROOF PROPOSITION 11. Membership in NP follows from the fact that we can easily verify a solution by calculating the average neighbor-to-neighbor relations in a labeled graph. We use a simple reduction from the problem of Graph 3-colorability. Graph 3-colorability is the question of whether there exists a labeling function $k : v \to \{1,2,3\}$ such that $k(u) \neq k(v)$ whenever $(u,v) \in E$ for a given graph $G = (V,E)$ and is well known to be NP-hard [50]. Assume now that we have a method that allows us to label any graph $G(V,E)$ following the heterophily matrix $\mathbf{H} = \frac{1}{2}\begin{bmatrix} 0 & 1 & 1 \\ 1 & 0 & 1 \\ 1 & 1 & 0 \end{bmatrix}$, i.e., neighboring nodes never have the same label. It follows immediately that such a solution would also be solution to graph 3-colorability. □

---

**Algorithm 1:** Planted graph generator that creates a synthetic directed graph with one planted edge potential.

**Input:** $(n, m, \alpha, \mathbf{H}, \text{dist})$
**Output:** $\mathbf{W}, \mathbf{X}$

1. Assign classes to nodes: $k(v), \mathbf{X}$
2. Calculate number of edges for each (type, type)-pair:
   $\mathbf{M} \in \mathbb{N}^{k \times k}, \mathbf{m}_{\text{out}}, \mathbf{m}_{\text{in}}$
3. Assign indegree and outdegree distributions to nodes: $d_{\text{out}}(v), d_{\text{in}}(v)$
4. Assign actual edges between nodes: $\mathbf{W}$

---

We thus need graph generators which generate both the graph topology (i.e., $\mathbf{W}$) and the node labels (i.e., $\mathbf{X}$) in the same process. We know of only two papers that have proposed graph generators that generate labeled data in the process: the early work by [32] and the very recent work by [49]. Neither graph generator is available. In addition, neither of the papers gives a way to know the "ground truth" actual potential matrix that was used to label data (e.g., [49] suggests this as future work).

We therefore had to implement our own synthetic graph generator with the additional design decision that any potential matrix can be "planted" as *exact graph property*. This allows us to separate the concern between (1) how does our method work on graphs with certain properties, (2) what is the variation in properties of a given generative model. By planting exact properties (instead of expected properties) we can focus on question (1) only. The random graph generator is described in detail in the next section. The standard stochastic blockmodel generates networks with a structure that is not similar to that found in most empirical network data.

## F. SYNTHETIC GRAPHS WITH A PLANTED DIRECTED COMPATIBILITY MATRIX

In the following, we describe a family of "*planted graph generators*" that create random synthetic graphs with a given set of parameters planted as exact graph properties.[14] Our motivation is to break apart the overall question of "relative performance between methods" into two distinct yet complementary questions: (*i*) How well do the methods perform *for a particular choice of parameters of a problem*? (*ii*) What parameters are representative of a particular domain? Our aim in this paper is to address only the former of the two questions, which can be answered by a completely controlled ground truth.

The idea is simple: we create *several random graphs with carefully chosen graph parameters* (or even exhaustively explore the entire parameter space) and then evaluate the *relative performance* of alternative methods as function of these parameters. Notice that we would not be able to systematically explore the differences with real-life data, as such data represent but a few particular choices of parameters. In contrast, our setup with carefully generated synthetic data gives us a methodical way to determine the conditions under which each method performs in relation to others.

### F.1 The graph generator (algorithm 1)

We next describe the planted graph generator that creates a random synthetic directed graph with a *single planted directed edge compatibility matrix* $\mathbf{H}$. This setup of one single directed potential corresponds to our formulation from [47, Corollary 17]. We are planting the compatibility by ensuring that the number of edges between nodes of different classes is proportional to the their compatibilities. In other words, $\mathbf{M} \propto \mathbf{H}$, where $M(j,i)$ is the number of directed edges from a node of class $j$ to a node of class $j$.

---
[14] Notice that "planted" here means that the resulting graphs have a desired graph property not just *in expectation*, but rather *exactly*.

The input to the algorithm is a tuple of parameters $(n, m, \alpha, \mathbf{H}, \text{dist})$ where $n$ is the number of nodes, $m$ is the number of edges, $\alpha$ is the node label distribution with $\alpha(i)$ being the fraction of nodes of class $i$ ($i \in [k]$), $\mathbf{H}$ is an arbitrary $k \times k$-edge potential, and dist is a family of degree distributions (e.g., uniform or power law with specified coefficient). The output is a directed graph with adjacency matrix $\mathbf{W}$ with $m = |\mathbf{W}|$ and explicit belief matrix $\mathbf{X}$ with each row $\mathbf{x}_v = \mathbf{X}(v,:)$ being an indicator vector that maps a node $v$ to its class $k(v)$. The graph generation proceeds in 4 steps (algorithm 1):

**Line 1:** The number of nodes for each class is given by: $\mathbf{n} = n\alpha$, with appropriate rounding. Each node gets assigned a random node class such that the number of nodes for each class is $n(i) = |\{v \mid k(v) = i\}|$. This can be done by random sampling *without* replacement, or simpler, by a random permutation of a label vector of size $n$. For example, assume $\mathbf{n} = [2, 3, 1]^\mathsf{T}$, we then simply use a random permutation of the tuple $(1, 1, 2, 2, 2, 3)$.

**Line 2:** Let $H_{\text{sum}} \triangleq \sum_{j,i} H(j,i)$ be the sum of all entries of the given potential $\mathbf{H}$ (notice that $H_{\text{sum}} = k^2$ in case $\mathbf{H}$ is centered around 1). Scaling $\mathbf{H}$ by $H_{\text{sum}}^{-1}$ leads to a "density matrix" that gives the relative fraction of edges between two classes among all edges. Multiplying by $m$ leads to $\mathbf{M} \triangleq \frac{m}{H_{\text{sum}}} \mathbf{H}$ where $M(j,i)$ is the number of edges in the graph from a node of class $j$ to a node of class $i$. The number of outgoing (or incoming) edges to nodes of a class is then the row-wise (or column-wise) sum: $m_{\text{out}}(j) = \sum_i M(j,i)$ (or $m_{\text{in}}(i) = \sum_j M(j,i)$). The average outdegree (or indegree) for each class is then: $\mathbf{d}_{\text{out}} = \mathbf{m}_{\text{out}} \oslash \mathbf{n}$ (or $\mathbf{d}_{\text{in}} = \mathbf{m}_{\text{in}} \oslash \mathbf{n}$).

**Line 3:** For each class $j$, we now have the number of nodes $n(j)$, the total number of outgoing edges $m_{\text{out}}(j)$ (and incoming edges $m_{\text{in}}(j)$), and a desired degree distribution family dist. From these numbers, we can now calculate an actual outdegree (and indegree) distribution. For example, assume that $n(j) = 3$, $m_{\text{out}}(j) = 11$, and dist is the family of power-law distributions with exponent $\beta = -1$.[15] Then the best fit actual degree distribution (i.e., the actual vector of integer outdegrees for that class) is $(6, 3, 2)$. Finally, the degrees are randomly assigned to each node per class. The result are two functions that assign outdegrees and indegrees *to each node*: $d_{\text{out}}(v)$ and $d_{\text{in}}(v)$. In our experiments, we assign indegree and outdegree independently. Alternatively, those could be made arbitrarily correlated (e.g., the node with highest indegree also has the highest outdegree).

**Line 4:** Thus far, we have: (1) for each node $v$, it's class $k(v)$, and the number of outgoing edges $d_{\text{out}}(v)$ and incoming edges $d_{\text{in}}(v)$; (2) for each class $j$, the row vector $\mathbf{M}(j,:)$ of the number of outgoing edges to all other classes, and the column vector $\mathbf{M}(:,j)$ of the number of incoming edges from all other classes. We next find a random edge assignment that fulfills these constraints. Our concrete instantiation works as follows: we initialize a counter matrix $\mathbf{M}'$ and counting vectors $\mathbf{d}'_{\text{out}}$ and $\mathbf{d}'_{\text{in}}$ with $\mathbf{M}$, $\mathbf{d}_{\text{out}}$, and $\mathbf{d}_{\text{in}}$, respectively. We then decrement $M(j,i)'$, $d'_{\text{out}}(s)$, and $d'_{\text{in}}(t)$ by one for any new edge that connects node $s$ with class $k(s) = j$ to node $t$ with class $k(t) = i$. We repeatedly pick an edge between classes from $\mathbf{M}'$ and find a random source and target consistent with $\mathbf{d}'_{\text{out}}$ and $\mathbf{d}'_{\text{in}}$, until all entries are 0. In this step, we may also add additional constraints when assigning new edges (e.g., that we never have a directed cycle of length 2, such as $s \to t \to s$).

Notice that steps 1-3 ensure that the resulting random graph has the desired parameters "planted" as exact graph properties. Here, we make the implicit assumption that higher order properties of a graph (e.g., the number of triangles in a graph) do not have significant impact on the *relative* performance between the methods we are comparing. One could think about refining step 4 in order to plant such triadic measures as well. For example, one could incorporate the idea of [30], which describes a preferential attachment graph generator with adjustable clustering coefficient, by constraining the set of edges from which additional edge is sampled.

EXAMPLE 12 (DIRECTED GRAPH 1). *Consider parameters $n = 1000$, $m = 3000$, $\alpha = [0.35, 0.25, 0.4]^\mathsf{T}$, $\mathbf{H} = \begin{bmatrix} 1 & h & 1 \\ 1 & 1 & h \\ h & 1 & 1 \end{bmatrix}$ with $h = 2$ or $h = 8$, and a power law distribution with exponent -0.5 as dist. Figure 12a shows the outdegree distribution among different classes of nodes. Figures 12b and 12c show the resulting "clustered adjacency matrices" for graphs with weak ($h = 2$) or strong potential ($h = 8$).*

## F.2 Parameter relationships

From the previous discussion follows that the average node degree per class is determined by the remaining parameters. Concretely the ratio between average outdegree of classes $j$ and $i$ is given by:

$$\frac{d_{\text{out}}(i)}{d_{\text{out}}(j)} = \frac{m_{\text{out}}(i)}{m_{\text{out}}(j)} \cdot \frac{n(j)}{n(i)}$$

$$= \frac{\sum_c \mathbf{H}(i,c)}{\sum_c \mathbf{H}(j,c)} \cdot \frac{\alpha(j)}{\alpha(i)} \quad (32)$$

In other words, one of the three parameters per class ($d_{\text{out}}(i)$, $\sum_c \mathbf{H}(i,c)$, $\alpha(i)$) needs to be unspecified. For example, if two classes have the same row sum in the compatibility matrix (and thus the same number of outgoing edges), then the outdegrees are inversely proportional to the number of nodes of that class. In turn, if we like to specify the node fractions, and have the same outdegrees, then the compatibility matrix needs to be adopted. We illustrate this relationship with two examples.

EXAMPLE 13 (DIRECTED GRAPH 2). *First, assume parameters $n = 120$, $m = 1080$, $\alpha = [\frac{1}{6}, \frac{1}{3}, \frac{1}{2}]$, $\mathbf{H} = \begin{bmatrix} 1 & 8 & 1 \\ 1 & 1 & 8 \\ 8 & 1 & 1 \end{bmatrix}$, and a powerlaw distribution with exponent $-0.4$. Figures 13a and 13d show the resulting degree distribution and clustered adjacency matrix. The latter represents the number of edges between different classes: $\mathbf{M} = \begin{bmatrix} 36 & 288 & 36 \\ 36 & 36 & 288 \\ 288 & 36 & 36 \end{bmatrix} = 36\mathbf{H}$. Notice that $|\mathbf{M}| = |\mathbf{W}| = 1080 = m = 9n$. Also notice that the vector of average outdegrees is $\mathbf{d}_{\text{Out}} = [18, 9, 6]$.*

EXAMPLE 14 (DIRECTED GRAPH 3). *Starting from example 13, we next assume that that all classes have the same average outdegree $d_{\text{Out}} = [9, 9, 9] = m/n$, yet they keep the uneven node distribution from before: $\alpha = [\frac{1}{6}, \frac{1}{3}, \frac{1}{2}]$. Based on eq. (32), we thus have to adjust the compatibility matrix by weighting the individual rows accordingly. We can do this, for example, by setting $\mathbf{H} = \begin{bmatrix} 1 & 8 & 1 \\ 2 & 2 & 16 \\ 24 & 3 & 3 \end{bmatrix}$. Figures 13b and 13e show the resulting degree distribution and clustered adjacency matrix with $\mathbf{M} = \begin{bmatrix} 18 & 144 & 18 \\ 36 & 36 & 288 \\ 432 & 54 & 54 \end{bmatrix} = 18\mathbf{H}$.*

---

[15]Power laws are varyingly defined in the literature in three different ways. In our formulation, we use the formulation in terms of "degree distribution": $d(v) \propto v^{-\delta}$ where $\delta(v)$ is the degree of the node with $v$-th highest degree. See [46, Sect. 4] for a detailed discussion on the relation between the three formulations.

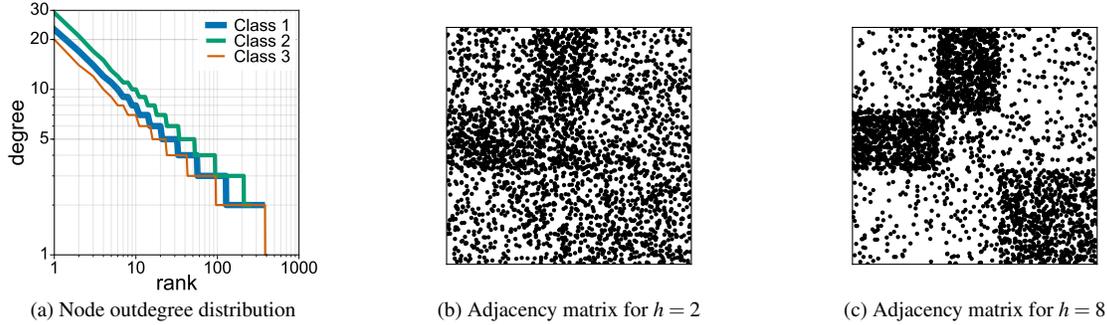

(a) Node outdegree distribution  (b) Adjacency matrix for $h = 2$  (c) Adjacency matrix for $h = 8$

Figure 12: Example 12: Our synthetic graph generator takes a tuple $(n, m, \alpha, \mathbf{H}, \mathsf{dist})$ and creates a random graph with those exact graph properties planted. For (b) and (c), node ids were permuted as to have nodes of identical classes with consecutive ids. We refer to this representation as the "clustered adjacency matrix".

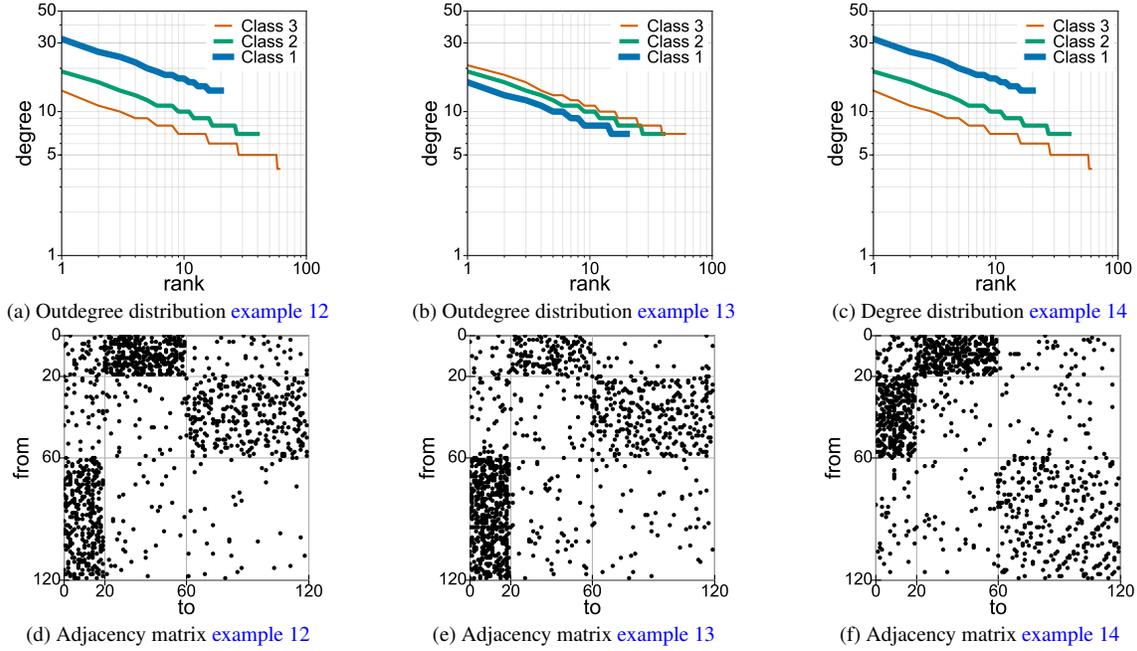

(a) Outdegree distribution example 12  (b) Outdegree distribution example 13  (c) Degree distribution example 14

(d) Adjacency matrix example 12  (e) Adjacency matrix example 13  (f) Adjacency matrix example 14

Figure 13: Examples 12 and 13: Directed graphs. Example 14: Undirected graph.

## F.3 Undirected graphs with a symmetric compatibility

We next discuss two simple modifications that allow algorithm 1 to create undirected graphs with a planted symmetric compatibility matrix.

1. In line 4, we only create a new edge from node $s$ to $t$, if the nodes are not yet connected in the reverse direction $t \to s$.
2. After the creation, we create for every edge $(s \to t)$, the corresponding backedge $(t \to s)$.

EXAMPLE 15 (UNDIRECTED GRAPH). *We use the same parameters as in example 13, but now create an undirected graph with $m = 540$ and $\mathbf{H} = \begin{bmatrix} 1 & 8 & 1 \\ 8 & 1 & 1 \\ 1 & 1 & 8 \end{bmatrix}$. Figures 13b and 13e show the resulting degree distribution and clustered adjacency matrix with $\mathbf{M} = \begin{bmatrix} 36 & 288 & 36 \\ 36 & 36 & 288 \\ 288 & 36 & 36 \end{bmatrix} = 18\mathbf{H}$. Notice that now $|\mathbf{M}| = |\mathbf{W}| = 1080 = 2m$.*

## Additional References